\documentclass{article}

\PassOptionsToPackage{numbers, compress}{natbib}

\usepackage{multirow}

\usepackage[final]{neurips_2023}

\usepackage[utf8]{inputenc} 
\usepackage[T1]{fontenc}    
\usepackage{url}            
\usepackage{booktabs}       
\usepackage{amsfonts}       
\usepackage{nicefrac}       
\usepackage{microtype}      
\usepackage{xcolor}         
\usepackage{graphicx}
\usepackage{subcaption}
\usepackage{wrapfig}
\usepackage{bbding}
\usepackage[colorlinks,linkcolor=blue]{hyperref}
\usepackage{ulem}
\usepackage{amsmath} 
\usepackage{longtable}
\usepackage{tabularx} 
\usepackage{lipsum}

\DeclareMathOperator*{\argmax}{arg\,max}  

\title{Large Language Models are Zero Shot Hypothesis Proposers}

\author{
    \textbf{Biqing Qi}$^{\,\ 1,2,3\,}$\thanks{~Equal contributions.}\\
    \texttt{qibiqing7@gmail.com}
    \And \textbf{Kaiyan Zhang}$^{\,1\ *}$\\
    \texttt{zhang-ky22@mails.tsinghua.edu.cn}
    \AND \textbf{Haoxiang Li}$^{\,1\,}$\\
    \texttt{hx-li20@mails.tsinghua.edu.cn}
    \And \textbf{Kai Tian}$^{\,1\,}$\\
    \texttt{tk23@mails.tsinghua.edu.cn}
    \And \textbf{Sihang Zeng}$^{\,4\,}$\\
    \texttt{zengsh@uw.edu}
    \And \textbf{Zhang-Ren Chen}$^{\,5\,}$\\
    \texttt{chenzhangren@ncu.edu.cn}\\
    \And \textbf{Jin-Fang Hu}$^{\,5\ \dag\,}$ \\
    \texttt{hujinfang333@126.com}
    \And \textbf{Bowen Zhou}$^{\, 1,2\,}$\thanks{~Corresponding author.}\\
    \texttt{zhoubowen@tsinghua.edu.cn}
    \AND
    $^1$ \normalfont Tsinghua University \quad 
    $^2$ Frontis.AI \quad
    $^3$ Harbin Institute of Technology\\
    $^4$ University of Washington \quad
    $^5$ The First Affiliated Hospital of Nanchang University \\
}

\begin{document}

\maketitle

\begin{abstract}
Significant scientific discoveries have driven the progress of human civilisation. The explosion of scientific literature and data has created information barriers across disciplines that have slowed the pace of scientific discovery. Large Language Models (LLMs) hold a wealth of global and interdisciplinary knowledge that promises to break down these information barriers and foster a new wave of scientific discovery. However, the potential of LLMs for scientific discovery has not been formally explored.
In this paper, we start from investigating whether LLMs can propose scientific hypotheses. To this end, we construct a dataset consist of background knowledge and hypothesis pairs from biomedical literature. 
The dataset is divided into training, seen, and unseen test sets based on the publication date to control visibility. 
We subsequently evaluate the hypothesis generation capabilities of various top-tier instructed models in zero-shot, few-shot, and fine-tuning settings, including both closed and open-source LLMs. 
Additionally, we introduce an LLM-based multi-agent cooperative framework with different role designs and external tools to enhance the capabilities related to generating hypotheses. 
We also design four metrics through a comprehensive review to evaluate the generated hypotheses for both ChatGPT-based and human evaluations.
Through experiments and analyses, we arrive at the following findings: 
1) LLMs surprisingly generate untrained yet validated hypotheses from testing literature. 
2) Increasing uncertainty facilitates candidate generation, potentially enhancing zero-shot hypothesis generation capabilities. 
These findings strongly support the potential of LLMs as catalysts for new scientific discoveries and guide further exploration.
\end{abstract}

\section{Introduction}

\begin{quote}
    \textit{``When nothing is sure, everything is possible.''} --- Margaret Drabble
\end{quote}

The pursuit of knowledge discovery stands as a cornerstone of human progress, driving innovation, and shaping our understanding of the world \cite{wang2023scientific,taylor2022galactica}. However, in recent times, the process of knowledge discovery has encountered formidable challenges, characterized by serendipity and sluggishness.
As the volume of data and literature continues to expand at an unprecedented rate, the ability to distill high-value insights and gain profound understanding from this wealth of information has become increasingly daunting\cite{taylor2022galactica}. Silos of information have erected themselves between disciplines, impeding the crucial cross-pollination of ideas and insights that could propel discovery to new heights.
Yet, amidst these challenges, there is a glimmer of hope. The advent of large-scale models (LLMs), possessing the capacity to harness a vast reservoir of world knowledge and span multiple domains, holds promise in revolutionizing the landscape of knowledge discovery. These models present an opportunity to break down the barriers between disciplines, enabling researchers to traverse the expansive sea of information with ease and efficiency.
Central to the process of knowledge discovery lies the formulation of sound hypotheses \cite{zhong2023goal,wang2023learning,boiko2023emergent,yang2023large}. However, a glaring gap persists in the arsenal of tools available to formally explore and evaluate hypotheses. While literature is replete with discussions on validation, it often overlooks the critical aspect of generating novel hypotheses.

In light of these challenges and opportunities, this paper delves into the current state of knowledge discovery, examining the hurdles posed by information explosion and disciplinary isolation. It explores the potential transformative role of LLMs in bridging these gaps, ultimately emphasizing the pivotal role of hypothesis generation in the knowledge discovery process. 
\begin{wrapfigure}{r}{0.5\textwidth} 
  \centering
  \includegraphics[width=0.5\textwidth]{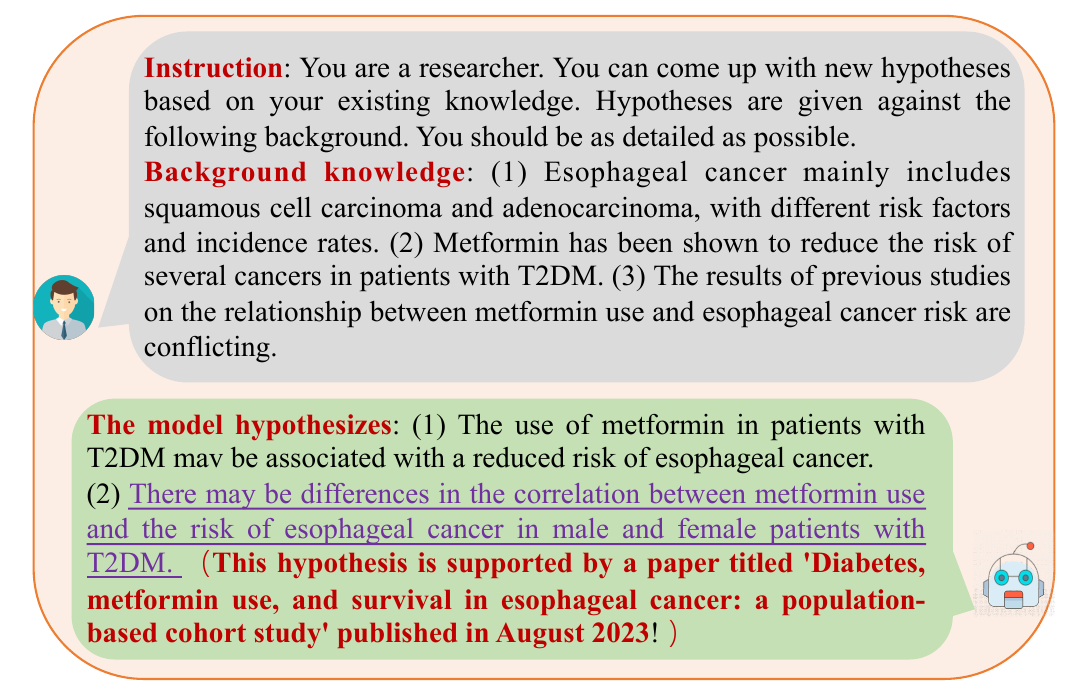} 
  \caption{Illustrating an generated hypothesis utilizing the fine-tuned 65B LLaMA within our constructed datasets, which closely match the findings in existing literature.} 
  \label{fig:example}
  \hfill
\end{wrapfigure}
Furthermore, it highlights the pressing need for tools and methodologies to facilitate hypothesis generation, thus propelling knowledge discovery into a new era of efficiency and innovation \cite{jain2023gflownets}.
Currently, both ChatGPT and GPT-4 undergo extensive pre-training on vast datasets and possess the capability of continuous updates. 
However, ensuring strict traceability of data sources becomes a challenging task, limiting our ability to explore zero-shot hypothesis generation.

The past literatures have explored scenarios of problem discovery, yet rigorous experimental designs to investigate whether LLMs can effectively propose genuine problems under zero-shot conditions remain lacking.
To tackle this issue, we assemble a dataset of biomedicine literature spanning from January 2000 to September 2023. 
This dataset is partitioned into training and testing sets, with the training set exclusively containing literature published before January 2023. 
We construct an unseen test set using literature from August 2023 and ensure that the evaluated LLMs have been trained on corpora before that date.
Additionally, we devise a multi-intelligent collaborative framework that incorporates search tools and role-playing to delve deeper into and uncover the potential for hypothesis generation.

Through experiments and analyses as shown in Figure~\ref{fig:example}, we draw the following findings: 1) LLMs surprisingly generate hypotheses that are untrained yet validated when tested against literature.
2) Increasing uncertainty levels can benefit by diversifying candidate generation and potentially enhancing zero-shot hypothesis generation capabilities. For instance, introducing heightened uncertainty through collaborative multi-agent approaches significantly improves the model's ability to generalize in zero-shot scenarios. However, integrating subsequent few-shot enhancements and using additional tools may reduce the model's proficiency in generating hypotheses. This phenomenon is likely due to the reduction of uncertainty, limiting the model's space for hypothesis generation. Consequently, it lacks consistent positive effects, underscoring the need for careful consideration of the type of external knowledge employed.
The above findings also support the notion: "\textit{\textbf{When nothing is sure, everything is possible}}."
Specifically, our contributions are as follows:

1) To rigorously validate the zero-shot and few-shot hypothesis generation potential of LLMs, we construct temporal biomedical instruction data and devised novel and effective experiments for comprehensive analysis and evaluation. To the best of our knowledge, this is the first work that formally designs experiments to investigate the zero shot hypothesis generation capacity of LLMs.

2) Through validation across different models and various scenario dimensions, we surprisingly find that LLMs possess rudimentary higher-order knowledge reasoning capabilities and can propose new hypothesis statements. This provides new empirical insights and pathways for knowledge discovery.

3) For a comprehensive review of the generated hypotheses, we design metrics across four dimensions for both ChatGPT-based and human evaluations. The correlation scores between ChatGPT evaluations and manual results indicate that LLMs also play a significant role in hypothesis evaluations.

4) To efficiently explore and further harness the capability of hypothesis generation, we introduce a multi-agent system based on LLMs. Through efficient collaboration among multiple models and tool utilization, we analyze the factors influencing hypothesis generation by LLMs.

\section{Process of Scientific Discovery}
\label{sec:process_scientific_discovery}
\begin{figure}[h] 
\centering 
\includegraphics[width=0.6\textwidth]{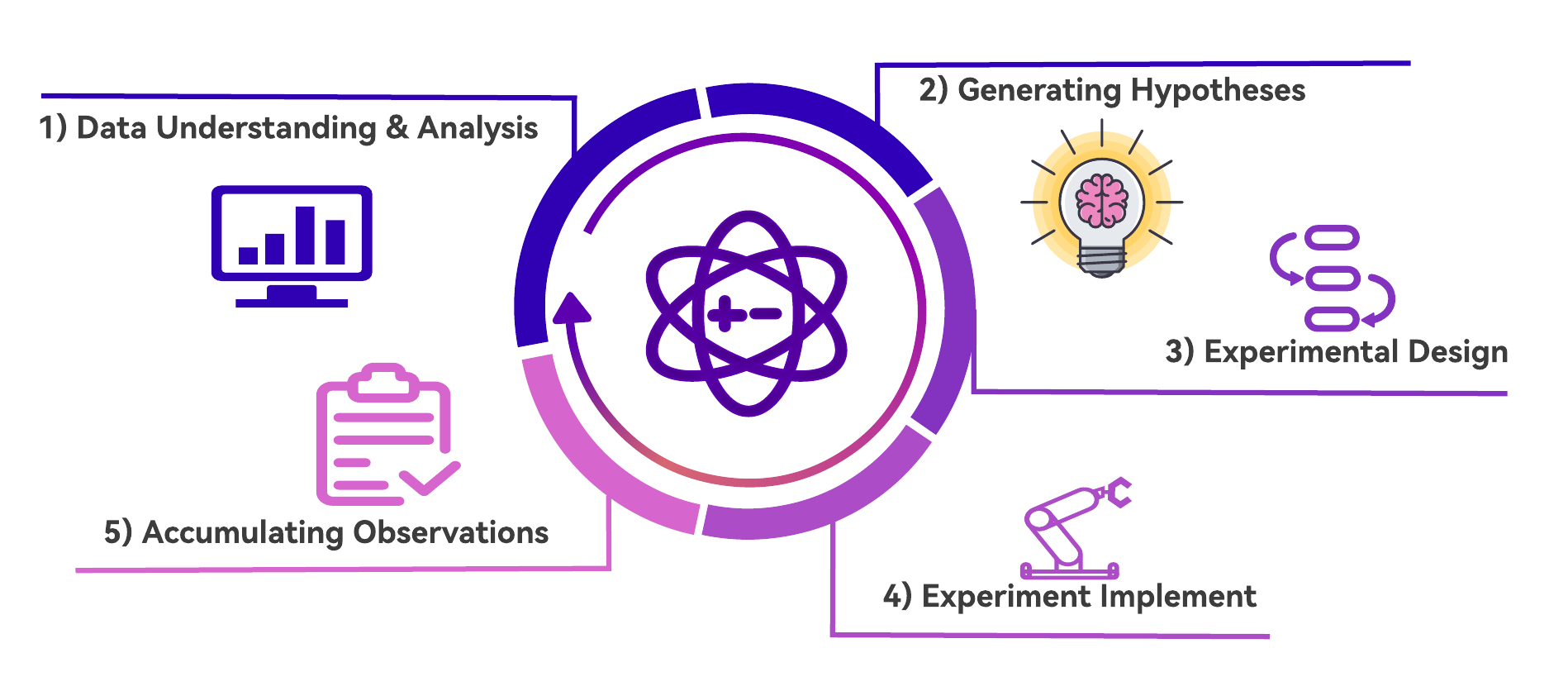} 
\caption{The iterative experimental loop of scientific discovery: observations and data accumulated from past experiments are analyzed and used to generate new hypotheses, and in turn new experiments that will yield new data to continue to cycle. In this paper, we mainly focus on investigating whether LLMs have the zero shot generalization ability to generate new hypotheses.} 
\label{fig:scientific_discovery} %
\end{figure}
Scientific discovery involves key components, each crucial for advancing our understanding of the natural world: data analysis, hypothesis formulation, experiment design, execution, and observation and reflection \cite{jain2023gflownets} as shown in Figure~\ref{fig:scientific_discovery}.

1) \textbf{Data Analysis}: Foundational in the scientific process, it entails collecting and examining data to discern patterns and anomalies, extracting insights through statistical techniques and visualization. It initiates scientific inquiry, guiding further exploration.
2) \textbf{Generating Hypotheses}: Among these components, hypothesis formulation is pivotal. It entails crafting informed guesses to explain observed phenomena. Hypotheses serve as guiding frameworks, directing and focusing research by articulating specific relationships and outcomes for experimental exploration.
3) \textbf{Experiment Design}: Once a hypothesis is set, designing experiments becomes essential to rigorously test its validity. This involves defining variables, specifying control groups, and outlining methods and procedures. Well-designed experiments ensure objective hypothesis testing and yield meaningful, informative results.
4) \textbf{Experiment Execution}: Meticulous execution of designed experiments and data collection are critical. Researchers adhere precisely to experimental protocols, recording observations, measurements, and unexpected findings. Integrity in execution ensures reliable, reproducible outcomes.
5) \textbf{Accumulating Observations}: After experiments, scientists engage in observation and reflection. They analyze collected data to determine if results support or refute the initial hypothesis. If unsupported, hypotheses may be revised or new ones formulated based on findings. Observation and reflection permit iterative refinement of scientific understanding.

\textbf{Hypothesis Pioneers Pathways: Guiding Knowledge Discovery}. While all components are essential, hypothesis formulation holds a unique position. It drives the scientific endeavor, guiding research question selection, experiment design, and data analysis. Well-constructed hypotheses not only provide direction but also lay the foundation for meaningful scientific discoveries by posing rigorously testable questions. Hypothesis formulation serves as the intellectual anchor steering scientific investigation and ultimately advancing knowledge.

\section{Can LLMs Truly Generate Zero-Shot Hypotheses?}
\label{sec:llm_sft}
In this section, we outline the methodology employed for a thorough assessment of LLMs' capacity to generate hypotheses under zero-shot conditions.
To accomplish this, we begin by defining the problem of hypothesis generation in zero-shot settings.
Next, we elucidate the process of dataset construction within the biomedical domain. 
Finally, we undertake comprehensive experiments to evaluate various instructed models across multiple dimensions, aiming to explore the factors influencing the ability of LLMs to propose improved hypotheses.

\subsection{Problem Definition}

Following the scientific discovery process outlined in Section~\ref{sec:process_scientific_discovery}, hypothesis generation typically occurs after thorough literature analysis and examination of specific phenomena.
To enhance evaluation effectiveness, we formalize this process as a text completion task. Given dataset {$D$}, an instruction $I$, and text pairs ${(X_i, Y_i)}^n_{i=1}$ containing background knowledge and corresponding hypotheses, extracted from medical papers, our objective is to assess model $M$ by having it generate hypotheses based on the task instruction and background knowledge, i.e., $M(I, X_i) = Y_i$, for each $i\in {1,...,n}$.
The objective function is formulated as:
$$
y^* = \argmax_{y_1, \ldots, y_n} \prod_{t=1}^{n} P(y_t | y_1, \ldots, y_{t-1}, I, X).
$$

\subsection{Dataset Construction}

\begin{wrapfigure}{r}{0.45\textwidth}
  \centering
  \includegraphics[width=0.45\textwidth]{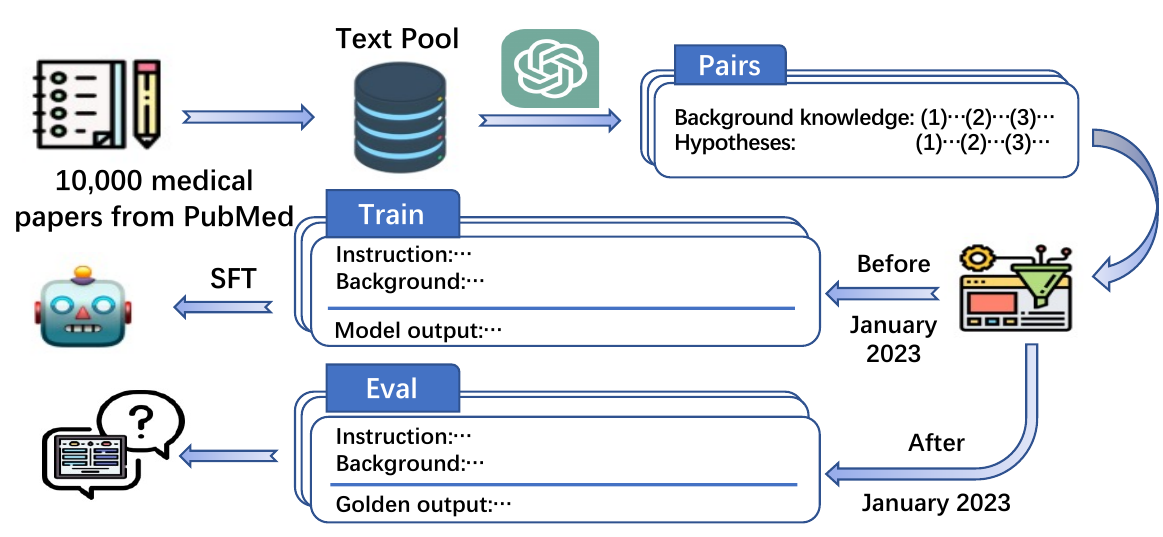} 
  \caption{Data partition pipeline.} 
  \label{fig:self_instruct}
\end{wrapfigure}

In this section, we detail the process of constructing datasets and ensuring the robustness of our evaluation. Prevalent LLMs, like Llama and ChatGPT, face challenges in tracing the origin of their knowledge due to continuous self-updating. To address this, we propose a novel approach to assess LLMs' hypothesis generation. Recognizing their potential impact on public domain data, we reconstruct a new biomedical literature dataset based on publication dates.

As depicted in Figure~\ref{fig:self_instruct}, we designated the year 2023 as the cut-off point. Our training dataset comprises literature published before January 2023, while the test dataset comprises literature published after January 2023, forming pairs of data with background knowledge and hypothesis proposals.
Due to the emergence of more advanced LLMs, our evaluations focus exclusively on the unseen test set, featuring literature published in August 2023. We selected instructed models fine-tuned before August 2023 for both evaluation and fine-tuning testing.
In our experimental setup, we implemented stringent measures to ensure the models had no prior exposure to the test data, affirming the validity of our experiments.
We strictly follow the standard pipeline as outlined in Self-Instruct~\cite{wang2022self} for our data generation process, encompassing four key steps:
1) Compose the paper set based on the topic and content of the literature.
2) Utilize chatgpt-turbo-3.5 to summarize the literature knowledge.
3) Generate background knowledge-assume pairs.
4) Filter low-quality data.
5) Split the dataset according to publication time.

\subsection{Datast Analysis}
In this section, we provide a comprehensive overview of the constructed dataset, encompassing details about the data acquisition strategy, dataset size, visibility control measures, distribution by year and month, as well as topic distribution.

We have created two datasets to maintain control over the visibility of hypotheses: 
1) \textbf{Seen dataset} This dataset comprises 2700 background and hypothesis pairs sourced from literature published before January 2023. This dataset was partitioned into training (2500) and validation (200) subsets (as well as seen test set). It is consistent with the corpus that the LLMs have been exposed to.
2) \textbf{Unseen dataset} The unseen dataset consists of 200 pairs extracted from papers published in August 2023, which the LLMs have not encountered during training and are used for testing purposes.

We also provide publication date and topic distribution of constructed dataset in Appendix~\ref{apx:details_dataset}.

\subsection{Experiment Setup}
\label{sec:experiment_setup}
In this section, we introduce experimental settings for hypothesis generation and evaluation.

\textbf{Models} For a fair comparison, we exclusively evaluate LLMs trained on corpora before March 2023 to ensure the test set remains unseen. We consider three categories of models in total:
1) API-based LLMs: this is mainly ChatGPT.
2) General domain instructed LLMs: These models consist of open-source models that have undergone fine-tuning based on Llama using general domain instructions. We primarily choose the top-tier models based on their performance rankings on the Alpaca Eval Leaderboard~\footnote{\url{https://tatsu-lab.github.io/alpaca_eval/}}.
3) Specific domain instructed LLMs: These include PMC-LLaMA~\cite{wu2023pmcllama}, and MedAlpaca~\cite{han2023medalpaca}. These models are trained on a variety of sources in medicine domain, such as medical books, PMC papers, medical dialogs, and others.
We provide detailed meta-information for various models, including their training data sources and publication dates, in Appendix~\ref{apx:details_models}.

\textbf{Prompts}
To ensure a consistent output format across different models, we create prompts in two formats: zero-shot and few-shot examples.
In our experiments, we adopt a 5-shot format, selecting examples from the training set before January 2023 using both randomly sampled and similarity retrieval methods.
We provide illustrations of zero-shot and few-shot prompts in Appendix~\ref{apx:prompt_design}.

\textbf{Finetuning}
To assess the hypothesis generation capability beyond zero-shot scenarios, we identify the top-performing open-source models through few-shot evaluation.
We proceed to fine-tune the full parameters of \verb|WizardLM-13B-V1.2| with the background and hypothesis pairs. The fine-tuning process consists of three epochs, employing a batch size of 8, a maximum sequence length of 2048 tokens, and a learning rate set at 3e-5.
We implement early stopping and select the best checkpoints based on their performance on the seen test dataset.

\textbf{Evaluation}
Given the disparities between the hypothesis generation task and traditional text generation tasks liking machine translation and summarization, with the former being more challenging and often involving uncertainty that extends beyond established ground truth, we approach our evaluation from two primary perspectives: conducting evaluations with and without golden hypotheses.
In evaluations with golden hypotheses, we employ standard text generation metrics, including BLEU and ROUGE in \verb|evaluate| library\footnote{\url{https://huggingface.co/docs/evaluate/index}}, to assess word overlap between the generated outputs and the ground truth.
The vastness of the hypothesis space renders it difficult to comprehensively assess the quality of generated hypotheses using word overlap metrics alone. 
To provide a more comprehensive evaluation of the generated hypotheses from multiple facets, we have thoughtfully devised four metrics: novelty, relevance, significance, and verifiability.
Inspired by recent research that highlights ChatGPT as proficient annotators~\cite{gilardi_chatgpt_2023, liu_g-eval_2023}, demonstrating a strong correlation with human ratings, we employ ChatGPT for further evaluation. 
In detail, we request ChatGPT to evaluate both the generated scientific hypotheses and the provided background across these aspects. The scoring scale ranges from 0 to 3, where a higher score indicates superior results. Additionally, we solicit ChatGPT to furnish a step-by-step explanation to substantiate the assigned score.
Moreover, we conduct human evaluation based on the four metrics for the top-tier models identified in the automatic evaluation in Section~\ref{sec:experiment_results}, and we provide a detailed description of this process in Section~\ref{sec:human_eval}.

\subsection{Experiment Results}
\label{sec:experiment_results}
\begin{table}[htbp]
  \caption{Results of various LLMs: We assess instructed models using zero-shot and few-shot format prompts to generate constrained outputs. 
  To provide a comprehensive assessment, we calculate the average scores for novelty, relevance, significance, and verifiability, denoted as Avg.
  Results marked with an asterisk (*) indicate that the few-shot prompts are constructed by retrieving samples from the training set that are similar to the background of inputs.
  To facilitate better comparison, we highlight the highest and sub-high score with both \textbf{bold} and \underline{underline} formatting under each category.
  }
  \label{tab:sft_result}
  \centering
  \scalebox{0.625}{
  \begin{tabular}{cllllllllll}
    \toprule
    \hline
    \multirow{ 2}{*}{\textbf{Category}} & \multirow{2}{*}{\textbf{Model}} &  \multicolumn{2}{c}{\textbf{Seen}} & \multicolumn{7}{c}{\textbf{Unseen}}  \\
    \cmidrule(r){3-4} \cmidrule(r){5-11} 
            &   & BLEU & ROUGE & BLEU & ROUGE & Novelty & Relevance & Significance & Verifiability & Avg \\
    \midrule
    \multirow{ 2}{*}{API-based}   
                   & gpt-3.5-turbo(0-shot) & 13.93 & 25.32  & 15.52 &  26.48  &  \textbf{1.42}  & \textbf{2.63} & 1.58 & 1.97 & 1.90  \\
                   & gpt-3.5-turbo(5-shot) & \underline{16.47} & \underline{27.07}  & \underline{16.49}  &  \underline{26.96}  &  \underline{1.22}  &  2.57  &  \underline{1.84}  &  \underline{2.03}  &  \underline{1.92} \\
                   & gpt-3.5-turbo(5-shot)* & \textbf{17.33}   &  \textbf{27.28}  &  \textbf{17.71}   &   \textbf{27.53}  &   1.02 & \underline{2.61} & \textbf{1.85}  & \textbf{2.36} & \textbf{1.96} \\ \midrule
    \multirow{ 15}{*}{General} 
           & Vicuna-33b-v1.3(0-shot)     & 13.97  & \underline{24.43} &   13.66  &  23.43   &  1.67  & 2.55  & 2.04  & 1.84 &  2.03 \\ 
           & Vicuna-33b-v1.3(5-shot)     & 11.23  & 22.54 &    11.49  &  22.68   &  1.60  & 2.40  & 1.67  & 1.90 &  1.89  \\ 
           & Vicuna-33b-v1.3(5-shot)*    & 12.78  & 24.11 &    13.12  &  23.66   &  1.19  & \underline{2.71}  & 2.00 &  2.17  &  2.02 \\ 
           & Llama-2-70b-chat(0-shot)    & 10.95  & 21.56 &  11.44  &  22.04   & \underline{1.86} &  2.41  & 1.91  & 1.98  &  2.04  \\
           & Llama-2-70b-chat(5-shot)    & 8.17   & 21.09 &   7.63  &  20.70   & \textbf{1.95} &  2.58  & 2.06  & \underline{2.22}  &  \textbf{2.20} \\
           & Llama-2-70b-chat(5-shot)*   & 8.40   & 21.65 &   9.66  &  22.43   &  1.43  &  2.50  & 1.94  & 2.15  &  2.01 \\
           & WizardLM-13B-V1.2(0-shot)   & 11.91  & 23.35 &  12.03  &  23.55   &  1.62  &  2.55  & 1.90  & 1.90  &  1.99 \\
           & WizardLM-13B-V1.2(5-shot)   & \underline{14.00}  & 24.30 &  \underline{13.82}  &  \underline{24.38}   &  1.33  & 2.54 &  1.81 &  \textbf{2.23} &  1.97 \\
           & WizardLM-13B-V1.2(5-shot)*  & 14.96  & 25.66 &  15.26  &  25.78 &  1.06  & 2.64  & 1.73  & 2.14 & 1.89 \\
           & WizardLM-70B-V1.0(0-shot)   & 13.45  & 24.12 &  \textbf{14.25}  &  \textbf{25.05}   &  1.57  &  2.45  & 1.74 & 1.89  & 1.91 \\
           & WizardLM-70B-V1.0(5-shot)   & \textbf{14.04} & \textbf{24.59} &  13.78  &  24.28   &  1.17  &  2.61 &  \textbf{2.12} &  2.14  &  2.01 \\
           & WizardLM-70B-V1.0(5-shot)*  & 14.46  & 24.78 &  15.26  &  25.56  &  0.97 & 2.67 & 1.85 & 1.99 & 1.87 \\
          & Openchat-v3.2-super(0-shot)  & 8.79   & 22.71 &  8.38   &  21.48  &  1.58 & 2.51 & 1.70 & 2.05 &  1.96 \\
          & Openchat-v3.2-super(5-shot)  & 12.46  & 23.60 &  12.58  &  24.21  &  1.06 & 2.64 & 2.09 & 2.20 &  2.00 \\
          & Openchat-v3.2-super(5-shot)* & 12.37  & 23.93 &  12.88  &  24.78  &  1.16 & \textbf{2.76} & \underline{2.10} &  \textbf{2.23} & \underline{2.07} \\\midrule
    \multirow{ 2}{*}{Medicine} 
           & MedAlpaca-13B(0-shot)   &  6.10  &  \underline{22.07}  &  5.82   &  \underline{20.49}   &  0.55  &  1.17  &  1.17  &  1.06  &  0.99 \\
           & MedAlpaca-13B(5-shot)   &  0.99  &   3.84  &  1.08   &   3.84   &  0.98  &  1.32  &  1.32  &  1.49  &  1.28 \\
           & MedAlpaca-13B(5-shot)*  &  4.60  &   9.36  &  4.50   &   9.07   &  1.09  &  1.40  &  1.20  &  \underline{1.53}  &  1.31 \\
           & PMC-LLaMA-13B(0-shot)   &  \textbf{22.89} &  \textbf{40.36}  &  \textbf{22.37}  &  \textbf{40.45}   &  0.76  &  \underline{1.94}  &  1.42  &  1.52  &  \underline{1.41} \\
           & PMC-LLaMA-13B(5-shot)   &  1.36  &   4.83  &  1.41   &   4.78   &  \underline{1.13}  &  1.45  &  \underline{1.36}  &  0.88  &  1.21 \\
           & PMC-LLaMA-13B(5-shot)*  &  \underline{6.21}  &  12.39  &  \underline{6.16}   &  12.13   &  \textbf{1.73}  &  \textbf{2.17}  &  \textbf{1.88}  &  \textbf{2.09}  &  \textbf{1.97} \\\midrule
    \multirow{1}{*}{SFT} 
           &  WizardLM-13B-V1.2   &  19.13 & 27.35 &  19.73  & 27.58  &  0.97  &  2.55  &  1.38 &  2.26 &  1.79 \\
    \hline
    \bottomrule
  \end{tabular}}
\end{table}

This section presents the results of hypothesis generation across various models in both zero-shot and few-shot settings. 
We primarily analyze the results from two perspectives: the impact of the zero-shot setting and the influence of introducing external knowledge on hypothesis generation.

\subsubsection{Impact of zero-shot settings}
The results presented in Table~\ref{tab:sft_result} demonstrate the significant impact of zero-shot settings in improving hypothesis generation, particularly in terms of fostering high novelty. We analyze these results from two key perspectives as following.

\textbf{Zero-shot Outperforms Few-shot.} Our findings indicate that, for extra large models like \verb|Llama-2-70b-chat| and \verb|WizardLM-70B-V1.0|, zero-shot performance surpasses that of the few-shot setting, where few-shot examples are obtained by randomly sampling. This suggests that the capacity of hypothesis generation is limited by the inclusion of few-shot examples, and models exhibit stronger abilities in a zero-shot setting. 

\textbf{Outperforming the Unseen Compared to the Seen Test Set.} Despite the visibility of literature published before 2022 in the pre-training corpus of most LLMs, we have categorized the test set into "seen" and "unseen." 
Typically, LLMs may excel in the "seen" test set due to the potential memorization of hypotheses present in the pre-training corpus, resulting in higher performance compared to the "unseen" test set. However, our results indicate that LLMs tend to perform better on the "unseen" test set. 
We speculate that this is because the complexity of hypothesis generation may hinder LLMs from effectively leveraging the dark knowledge in their parameters.

\subsubsection{Influence of external knowledge}
Based on the results, we observe that the introduction of external knowledge, such as few-shot examples, domain adaptation, and instruction fine-tuning, does not consistently enhance the ability of hypothesis proposing.

\textbf{Few-Shot Examples Enhance Verifiability but Decrease Novelty.} 
In comparison to zero-shot settings, models using few-shot prompts benefit from the provided examples, resulting in very high matching rates. 
Regarding word overlap metrics, including BLEU and ROUGE, most models, especially \verb|WizardLM| series models, and \verb|Openchat-v3.2-super|, show improved performance when provided with in-context examples, with retrieved examples being particularly beneficial for their generations.
However, it's important to note that these few-shot prompts significantly increase verifiability while simultaneously leading to lower levels of novelty compared to zero-shot results.

\textbf{Randomly Sampled Few-Shot Examples vs. Similarity Retrieval.}
Given that randomly sampled in-context examples often differ from the provided background in terms of topics or domains, this can potentially confuse LLMs and lead to decreased performance. 
In our pursuit of further exploration into the hypothesis generation capabilities of LLMs, we retrieve examples from the training dataset based on their similarity to the given background. 
The results indicate that similarity retrieval can further enhance performance.

\textbf{Instruction Tuning Enhances LLM Performance.} 
Following fine-tuning on a dataset comprising background and hypothesis pairs, \verb|WizardLM-13B-V1.2| attains superior performance, surpassing even \verb|gpt-3.5-turbo| and \verb|WizardLM-70B-V1.0|. This finding underscores that domain adaptation remains a valuable approach to enhance the hypothesis generation capabilities of LLMs. It not only leads to greater resource efficiency but also supports the promotion of privacy in a localized context.

\textbf{Impact of Domain Adaptation on Hypothesis Generation.} We have also conducted an analysis of the influence of fine-tuning for domain adaptation on hypothesis generation. In this comparison, we utilize instructed models adapted to the field of medicine.
The results obtained from \verb|MedAplaca| and \verb|PMC-LLaMA| indicate that domain adaptation can significantly improve word overlap performance. However, the metrics derived from ChatGPT suggest that domain adaptation has only a minimal effect on hypothesis generation. This discrepancy between word overlap metrics and ChatGPT's evaluation highlights the need for more comprehensive and unified metrics in the context of hypothesis generation tasks.

\subsection{Human Evaluation and Case Study}
\label{sec:human_eval}
In this section, we conduct a human evaluation to assess the generated hypotheses and calculate coherence scores to compare them with ChatGPT evaluation scores, guiding further evaluation efforts.

\subsubsection{Evaluation Settings}
\textbf{Evaluation Metrics} 
To comprehensively evaluate the generations manually and simultaneously assess the quality of ChatGPT's evaluations, we continue to utilize the four metrics outlined in Section~\ref{sec:experiment_setup}, which encompass novelty, relevance, significance, and verifiability.
The range of each metric remains from 0 to 3, with higher values indicating better performance.
Additionally, we calculate coherence scores between human evaluations and ChatGPT evaluations. 

\textbf{Selection of Models}
Given the constraints associated with the cost of human evaluation, our primary objective is to assess whether LLMs can produce valuable hypotheses, rather than striving for state-of-the-art performance. As a result, we exclusively perform human evaluation on the outputs generated by the LLM that ranks highest in performance based on automatic metrics and ChatGPT evaluation.
Furthermore, we aim to encompass a variety of prompts and models in our evaluation. The final models selected for human evaluation are detailed in Table~\ref{tab:human_eval}.

\textbf{Evaluation Details} We randomly selected 100 examples from the unseen test set and had three evaluators with a biomedical background assign scores for each metric to each model.

\subsubsection{Evaluation Results}
As depicted in Table~\ref{tab:human_eval}, the human evaluations exhibit a strong correlation with ChatGPT's evaluations, with Pearson and Spearman coefficients exceeding 0.7 for all models. These results strongly support our earlier findings regarding the influence of zero-shot learning and external knowledge. This reinforces our assertion that large language models can effectively propose hypotheses and significantly contribute to scientific discovery.
For additional insights, we present correlation coefficients between word overlap scores and manual scores in the Appendix~\ref{apx:additional_results}, revealing lower coherence and highlighting the need for more advanced evaluation metrics.

We also conduct a case study that showcases the hypotheses generated by various models and includes examples of step-by-step evaluations by ChatGPT. Details can be found in Appendix~\ref{apx:case_study}.

\begin{table}[htbp]
  \small
  \caption{
  This table presents the results of human evaluation. The \textit{Avg Coefficient} are used to assess the correlation between the average scores obtained from ChatGPT and those from human evaluation.
  }
  \label{tab:human_eval}
  \centering
  \scalebox{0.70}{
  \begin{tabular}{clllllllll}
    \toprule
    \hline
     \multirow{2}{*}{Category} & \multirow{2}{*}{Model} & ChatGPT & \multicolumn{5}{c}{Human Eval}  & \multicolumn{2}{c}{Avg Coefficient} \\
    \cmidrule(lr){4-8} \cmidrule(lr){9-10} 
     & &  Eval.Avg  &  Novelty & Relevance & Significance & Verifiability & Avg & Pearson & Spearman \\
    \midrule
     \multirow{2}{*}{API-based} 
     & gpt-3.5-turbo(0-shot)    &  1.90  &  1.54  &  2.69  &  1.77  &  2.08  &  2.02  &  0.87  &  0.78 \\
     & gpt-3.5-turbo(5-shot)*   &  1.96  &  1.31  &  2.62  &  2.08  &  2.62  &  2.15  &  0.80  &  0.78 \\
     \midrule
     \multirow{5}{*}{General} 
     & Llama-2-70b-chat(0-shot) &  2.04  &  1.77  &  2.23  &  1.92  &  1.92  &  1.96  &  0.89  &  0.84 \\
     & Llama-2-70b-chat(5-shot) &  2.20  &  2.15  &  2.77  &  2.08  &  2.31  &  2.33  &  0.96  &  0.90 \\
     & Llama-2-70b-chat(5-shot)*&  2.01  &  1.38  &  2.62  &  2.31  &  2.00  &  2.08  &  0.97  &  0.94 \\
     & WizardLM-70B-V1.0(0-shot)&  1.91  &  1.38  &  2.31  &  1.54  &  2.00  &  1.81  &  0.90  &  0.75 \\
     & WizardLM-70B-V1.0(5-shot)&  2.01  &  1.15  &  2.69  &  2.46  &  1.77  &  2.02  &  0.85  &  0.89 \\
     \midrule
     \multirow{2}{*}{Medicine} 
     & PMC-LLaMA-13B(0-shot)    &  1.41  &  1.00  &  2.62  &  1.92  &  2.00  &  1.88  &  0.73  &  0.73 \\
     & PMC-LLaMA-13B(5-shot)*   &  1.97  &  1.85  &  2.23  &  1.92  &  1.69  &  1.92  &  0.95  &  0.94 \\
     \midrule
     SFT & WizardLM-13BV1.2     &  1.79  &  0.85  &  2.77  &  1.23  &  2.23  &  1.77  &  0.83  &  0.85 \\
     \hline
    \bottomrule
  \end{tabular}}
\end{table}

\section{
Can agent collaboration enhance LLMs' zero-shot generalization?
}

\label{sec:multi_agents}
In this section, we will strive to enhance the ability of LLMs in hypothesis generation through multi-agent collaboration and the use of tools.
Our objective is to improve hypothesis efficiency by employing multi-agent collaboration, simulating real-world research scenarios. 
To begin, we introduce the conceptual system of multi-agent collaboration for hypothesis generation, drawing inspiration from scientific research. Subsequently, we present the role design and the tools use in this context. 
Finally, we present preliminary validated results of the multi-agent system using our proposed BHP dataset.

\subsection{Multi-agent Framework}
\begin{figure}[h]
  \centering
  \includegraphics[width=0.75\textwidth]{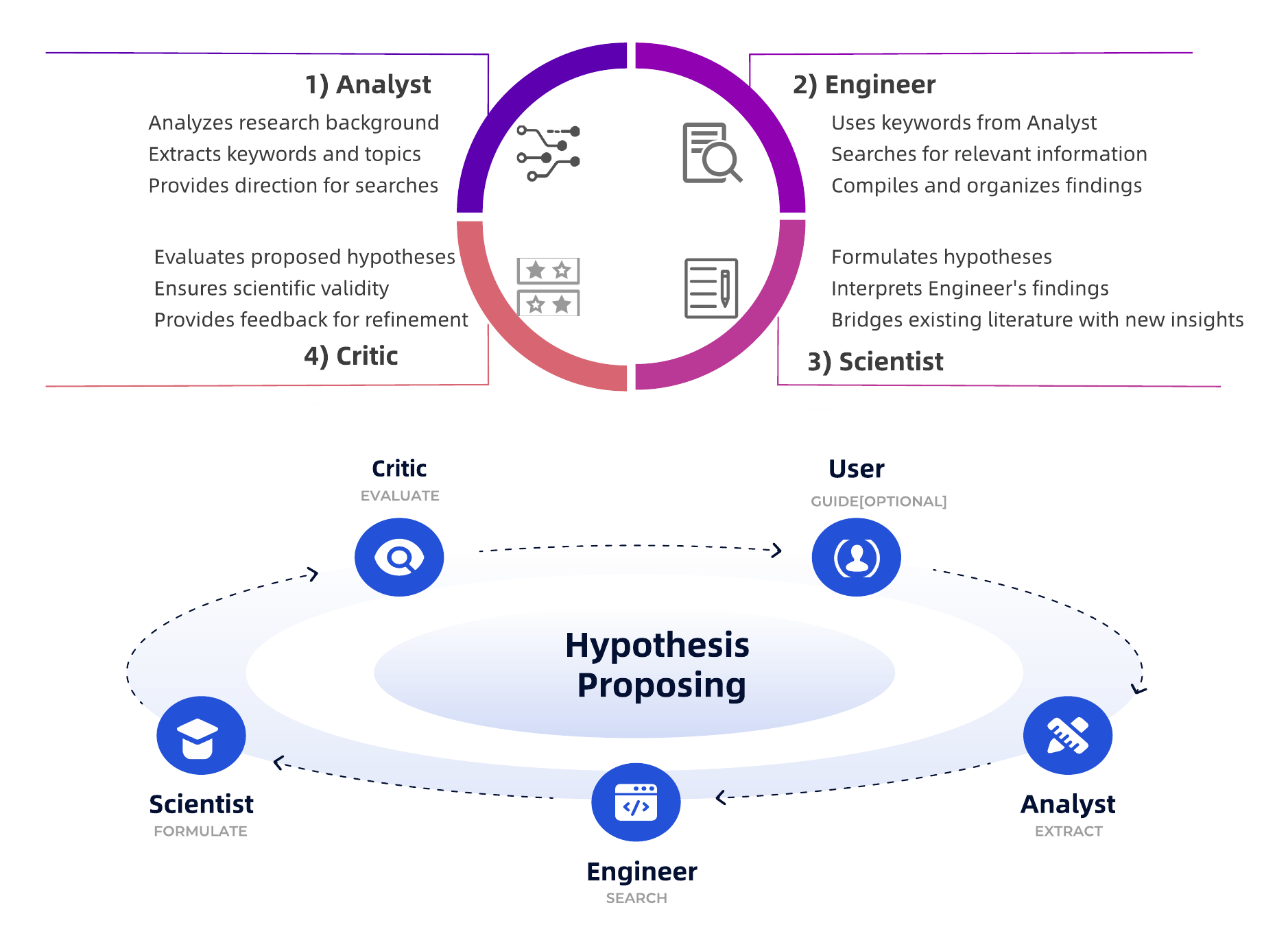} 
 \caption{The conceptual system of multi-agent collaboration for hypothesis generation. The overall prototyping process is illustrated below, allowing users to choose optional involvement. We offer core role descriptions of multi-agents and the fully automated system above.} 
  \label{fig:multi_agent}
\end{figure}

Inspired by the structured methodology detailed in Section~\ref{sec:process_scientific_discovery}, we introduce a comprehensive framework tailored for hypothesis formulation. This framework encapsulates a multi-agent system where each agent assumes a distinct role, mirroring the collaborative nature of scientific endeavors. Through a symbiotic and iterative process, these agents collaborate to craft hypotheses that are not only grounded in existing knowledge but also pave the way for novel insights. By emulating the essence of scientific discovery, our framework strives to produce hypotheses that are both innovative and scientifically robust. As depicted in Figure~\ref{fig:multi_agent}, we have partitioned the framework into five components, encompassing four automated agents and the option for human involvement within the loop. 

\textbf{Role Design } In our proposed multi-agent framework, each component plays a distinct and pivotal role.
The \textbf{Analyst} serves as the foundation, meticulously extracting and defining core elements from the research background. Its primary objective is to interpret the literature, distilling it into keywords or topics that subsequently guide the Engineer's search efforts. 
The \textbf{Engineer}, leveraging these keywords, embarks on a mission to retrieve and organize pertinent information. They meticulously plan and execute detailed searches, ensuring that the findings are compiled in a structured manner. 
This organized materials then lands in the domain of the \textbf{Scientist}, whose objective is to weave together the Engineer's findings with the original research background. Through careful interpretation, the Scientist crafts a hypothesis that is both grounded in existing knowledge and offers a fresh perspective. 
However, before this hypothesis is finalized, it undergoes scrutiny by the \textbf{Critic}. The Critic's role is paramount in ensuring the hypothesis's robustness, coherence, and novelty. They evaluate the hypothesis against the backdrop of the research background, ensuring it stands up to academic rigor. Feedback from the Critic, if necessary, loops back to refine the hypothesis or prompts the Analyst for further insights, creating a cyclical and iterative process of refinement.

\textbf{Tool Use } To explore external knowledge beyond the inherent dark knowledge within LLMs, we integrate the Engineer agent with search engines, mainly PubMed~\footnote{\url{https://pubmed.ncbi.nlm.nih.gov/}}. 
Similarly, to control the visibility of the unseen test dataset, we filter and exclude literature published after January 2023 from the search results. 
We carry out tool use experiments using ReAct~\cite{yao_react_2023} and OpenAI function calling. ReAct is a method that extends the concept of Chain of Thought (CoT)~\cite{wei2022chain}, involving thinking before taking action and subsequently making observations based on feedback from the environment. 
In our experiments, we instruct the LLMs to initially contemplate the provided background information and then make a decision regarding whether to utilize tools. 
Upon receiving feedback from the tools, the LLMs are expected to identify supporting evidence in the results or potentially make further tool requests. 
The LLMs are responsible for concluding the hypothesis generation process and summarizing the hypotheses independently.
In the case of OpenAI function calling, we directly specify tools for publication searching and transmit them to OpenAI APIs. 
This process is roughly implemented through fine-tuning, as described in ToolFormer~\cite{schick_toolformer_2023}.

\subsection{Experiment Results}

Our primary focus is to investigate the impact of tool use and multi-agent collaboration on hypothesis generation. We present the experimental results in Table~\ref{tab:exp_agent}.
Based on the results, we summarize our findings from two perspectives: tool use and role-playing.

\textbf{Influence of Tool Use}  Based on our results, we observe that tool use has minimal impact on improving the hypothesis generation ability of LLMs. 
This observation aligns with the findings presented in the previous sections regarding the analysis of external knowledge.
Notably, the ReAct-based method performs worse than OpenAI function calling. 
It is also evident that LLMs struggle to identify useful information and exhibit weaknesses in the thought-action-observation process, even when utilizing the official interface from OpenAI. 
Hypothesis generation is indeed a challenging task that necessitates iterative discussions and the exchange of ideas among various individuals.

\textbf{Multi-Agent Collaboration} 
In addition to tool use, our findings suggest that the division of labor and interaction among multi-agents can significantly enhance the model's capability to propose hypotheses by introducing uncertainty.
This mirrors the dynamics of real-world scientific research, where hypotheses are formulated through iterative discussions and refutations. Additionally, it is worth noting that tool use can further enhance the performance of the multi-agent framework.

\begin{table}[h]
  \caption{Results of individual agents and multi-agent systems, both with and without the use of tools, on the unseen test dataset. The results demonstrate that both multi-agent systems and the utilization of tools enhance the ability of LLMs in hypothesis generation. Among the various types of models, both $2^a$ and $2^b$ are evaluated with tool use. The difference between them lies in their implementations: ReAct~\cite{yao_react_2023} and OpenAI function calling~\protect\footnotemark, respectively. 
  }
  \label{tab:exp_agent}
  \centering
  \scalebox{0.75}{
  \begin{tabular}{clllllllll}
    \toprule
    \hline
    \multirow{2}{*}{Model} & \multicolumn{2}{c}{Influence Factor} & \multicolumn{2}{c}{Automatic} & \multicolumn{5}{c}{GPT-4 Evaluation}  \\
    \cmidrule(lr){2-3} \cmidrule(lr){4-5} \cmidrule(lr){6-10}
            & Multi-agent&   Tool use  &   BLUE   & ROUGE  & Novelty & Relevance & Significance & Verifiability & Avg \\
    \midrule
     $1$    &   -        & -           &   \underline{15.52}  &  \textbf{26.48}  & 1.23 & 2.57 & 1.84 & 2.03 & 1.92 \\
     $2^a$  &   -        & \Checkmark  &   14.94  &  24.16  & 0.78 & 2.42 & 1.18 & 1.87 & 1.56 \\
     $2^b$  &   -        & \Checkmark  &   \textbf{15.87}  &  \underline{24.94}  & 0.57 & 2.58 & 0.93 & 1.89 & 1.49 \\
     $3$   & \Checkmark & -           &   11.71  &  22.11  & \underline{1.35} & \textbf{2.85} & \underline{2.05} & \textbf{2.10} & \textbf{2.09} \\
     $4$   & \Checkmark & \Checkmark  &   11.18  &  22.04  & \textbf{1.52} &  \underline{2.66}  &  \textbf{2.06}  &  \underline{2.05}  &  \underline{2.07}  \\
     \hline
    \bottomrule
  \end{tabular}}
\end{table}
\footnotetext{\url{https://openai.com/blog/function-calling-and-other-api-updates}}

\section{Conclusion}
From the hypothesis-proposer perspective, we investigated LLMs' zero-shot generalisation ability in scientific research. Specifically, we first build a comprehensive corpus based on biomedical literature, split by publication date, including background knowledge and hypothesis pairs.
This corpus is then used as a basis for fine-tuning LLMs, leading to the generation of the LLM.
To further analysis and enhance the capabilities of the hypothesis proposer, we introduce a LLM-based multi-agent collaboration system.
Experimental results show that fine-tuned LLMs of various sizes can propose new hypotheses that did not appear in the training data but can be confirmed by the test literature, with performance comparable to ChatGPT and in some cases even better. 
Notably, our study revealed that introducing uncertainty into processes and operations enhances zero-shot generalization capabilities. 
These findings confirm the potential of LLMs to propose new hypotheses and offers hope for future unlocked scientific discovery.
In future work, we will focus on optimizing models and generating hypotheses guided by effective uncertainty assessment metrics.

\section*{Acknowledgements}
We extend our gratitude to the anonymous reviewers for their insightful feedback.

\bibliographystyle{plain}
\bibliography{neurips_2023}

\begin{thebibliography}{10}

\bibitem{boiko2023emergent}
Daniil~A Boiko, Robert MacKnight, and Gabe Gomes.
\newblock Emergent autonomous scientific research capabilities of large
  language models.
\newblock {\em arXiv preprint arXiv:2304.05332}, 2023.

\bibitem{bran_chemcrow_2023}
Andres~M. Bran, Sam Cox, Andrew~D. White, and Philippe Schwaller.
\newblock {ChemCrow}: {Augmenting} large-language models with chemistry tools,
  June 2023.
\newblock arXiv:2304.05376.

\bibitem{cao_life_2023}
Boxi Cao, Hongyu Lin, Xianpei Han, and Le~Sun.
\newblock The {Life} {Cycle} of {Knowledge} in {Big} {Language} {Models}: {A}
  {Survey}, March 2023.
\newblock arXiv:2303.07616 [cs].

\bibitem{changNewMechanicalMarkers2023}
Zhuo Chang, Jing Zhang, Yilun Liu, Huajian Gao, and Guang-Kui Xu.
\newblock New {{Mechanical Markers}} for {{Tracking}} the {{Progression}} of
  {{Myocardial Infarction}}.
\newblock {\em Nano Letters}, 23(16):7350--7357, August 2023.

\bibitem{chen_asking_2023}
Xiaoyu Chen, Shenao Zhang, Pushi Zhang, Li~Zhao, and Jianyu Chen.
\newblock Asking {Before} {Action}: {Gather} {Information} in {Embodied}
  {Decision} {Making} with {Language} {Models}, May 2023.
\newblock arXiv:2305.15695 [cs].

\bibitem{chen_empowering_2023}
Yulin Chen, Ning Ding, Hai-Tao Zheng, Zhiyuan Liu, Maosong Sun, and Bowen Zhou.
\newblock Empowering {Private} {Tutoring} by {Chaining} {Large} {Language}
  {Models}, September 2023.
\newblock arXiv:2309.08112 [cs] version: 1.

\bibitem{fu_improving_2023}
Yao Fu, Hao Peng, Tushar Khot, and Mirella Lapata.
\newblock Improving {Language} {Model} {Negotiation} with {Self}-{Play} and
  {In}-{Context} {Learning} from {AI} {Feedback}, May 2023.
\newblock arXiv:2305.10142 [cs].

\bibitem{gilardi_chatgpt_2023}
Fabrizio Gilardi, Meysam Alizadeh, and Maël Kubli.
\newblock {ChatGPT} {Outperforms} {Crowd}-{Workers} for {Text}-{Annotation}
  {Tasks}.
\newblock {\em Proceedings of the National Academy of Sciences},
  120(30):e2305016120, July 2023.
\newblock arXiv:2303.15056 [cs].

\bibitem{gong_mindagent_2023}
Ran Gong, Qiuyuan Huang, Xiaojian Ma, Hoi Vo, Zane Durante, Yusuke Noda, Zilong
  Zheng, Song-Chun Zhu, Demetri Terzopoulos, Li~Fei-Fei, and Jianfeng Gao.
\newblock {MindAgent}: {Emergent} {Gaming} {Interaction}, September 2023.
\newblock arXiv:2309.09971 [cs] version: 1.

\bibitem{han2023medalpaca}
Tianyu Han, Lisa~C Adams, Jens-Michalis Papaioannou, Paul Grundmann, Tom
  Oberhauser, Alexander L{\"o}ser, Daniel Truhn, and Keno~K Bressem.
\newblock Medalpaca--an open-source collection of medical conversational ai
  models and training data.
\newblock {\em arXiv preprint arXiv:2304.08247}, 2023.

\bibitem{hong_metagpt_2023}
Sirui Hong, Xiawu Zheng, Jonathan Chen, Yuheng Cheng, Jinlin Wang, Ceyao Zhang,
  Zili Wang, Steven Ka~Shing Yau, Zijuan Lin, Liyang Zhou, Chenyu Ran, Lingfeng
  Xiao, and Chenglin Wu.
\newblock {MetaGPT}: {Meta} {Programming} for {Multi}-{Agent} {Collaborative}
  {Framework}, August 2023.
\newblock arXiv:2308.00352 [cs].

\bibitem{hu_chatdb_2023}
Chenxu Hu, Jie Fu, Chenzhuang Du, Simian Luo, Junbo Zhao, and Hang Zhao.
\newblock {ChatDB}: {Augmenting} {LLMs} with {Databases} as {Their} {Symbolic}
  {Memory}, June 2023.
\newblock arXiv:2306.03901 [cs].

\bibitem{jain2023gflownets}
Moksh Jain, Tristan Deleu, Jason Hartford, Cheng-Hao Liu, Alex
  Hernandez-Garcia, and Yoshua Bengio.
\newblock Gflownets for ai-driven scientific discovery.
\newblock {\em Digital Discovery}, 2(3):557--577, 2023.

\bibitem{kang_think_2023}
Jikun Kang, Romain Laroche, Xindi Yuan, Adam Trischler, Xue Liu, and Jie Fu.
\newblock Think {Before} {You} {Act}: {Decision} {Transformers} with {Internal}
  {Working} {Memory}, May 2023.
\newblock 0 citations (Semantic Scholar/arXiv) [2023-05-30] arXiv:2305.16338
  [cs].

\bibitem{li_camel_2023}
Guohao Li, Hasan Abed Al~Kader Hammoud, Hani Itani, Dmitrii Khizbullin, and
  Bernard Ghanem.
\newblock {CAMEL}: {Communicative} {Agents} for "{Mind}" {Exploration} of
  {Large} {Scale} {Language} {Model} {Society}, March 2023.
\newblock arXiv:2303.17760 [cs].

\bibitem{liu_g-eval_2023}
Yang Liu, Dan Iter, Yichong Xu, Shuohang Wang, Ruochen Xu, and Chenguang Zhu.
\newblock G-{Eval}: {NLG} {Evaluation} using {GPT}-4 with {Better} {Human}
  {Alignment}, May 2023.
\newblock arXiv:2303.16634 [cs].

\bibitem{maas_show-1_nodate}
Philipp Maas, Frank Carey, Chris Wheeler, Edward Saatchi, Pete Billington, and
  Jessica~Yaffa Shamash.
\newblock {SHOW}-1 and {Showrunner} {Agents} in {Multi}-{Agent} {Simulations}.
\newblock {\em arXiv preprint}, 2023.

\bibitem{nakano_webgpt_2022}
Reiichiro Nakano, Jacob Hilton, Suchir Balaji, Jeff Wu, Long Ouyang, Christina
  Kim, Christopher Hesse, Shantanu Jain, Vineet Kosaraju, William Saunders,
  Xu~Jiang, Karl Cobbe, Tyna Eloundou, Gretchen Krueger, Kevin Button, Matthew
  Knight, Benjamin Chess, and John Schulman.
\newblock {WebGPT}: {Browser}-assisted question-answering with human feedback,
  June 2022.
\newblock arXiv:2112.09332.

\bibitem{parisi_talm_2022}
Aaron Parisi, Yao Zhao, and Noah Fiedel.
\newblock {TALM}: {Tool} {Augmented} {Language} {Models}, May 2022.
\newblock arXiv:2205.12255 [cs].

\bibitem{park_generative_2023}
Joon~Sung Park, Joseph~C. O'Brien, Carrie~J. Cai, Meredith~Ringel Morris, Percy
  Liang, and Michael~S. Bernstein.
\newblock Generative {Agents}: {Interactive} {Simulacra} of {Human} {Behavior},
  August 2023.
\newblock arXiv:2304.03442 [cs].

\bibitem{patil_gorilla_2023}
Shishir~G. Patil, Tianjun Zhang, Xin Wang, and Joseph~E. Gonzalez.
\newblock Gorilla: {Large} {Language} {Model} {Connected} with {Massive}
  {APIs}, May 2023.
\newblock arXiv:2305.15334 [cs].

\bibitem{qian_communicative_2023}
Chen Qian, Xin Cong, Wei Liu, Cheng Yang, Weize Chen, Yusheng Su, Yufan Dang,
  Jiahao Li, Juyuan Xu, Dahai Li, Zhiyuan Liu, and Maosong Sun.
\newblock Communicative {Agents} for {Software} {Development}, August 2023.
\newblock arXiv:2307.07924 [cs].

\bibitem{rawte_survey_2023}
Vipula Rawte, Amit Sheth, and Amitava Das.
\newblock A {Survey} of {Hallucination} in {Large} {Foundation} {Models},
  September 2023.
\newblock arXiv:2309.05922 [cs].

\bibitem{schick_toolformer_2023}
Timo Schick, Jane Dwivedi-Yu, Roberto Dessì, Roberta Raileanu, Maria Lomeli,
  Luke Zettlemoyer, Nicola Cancedda, and Thomas Scialom.
\newblock Toolformer: {Language} {Models} {Can} {Teach} {Themselves} to {Use}
  {Tools}, February 2023.
\newblock arXiv:2302.04761.

\bibitem{shen_hugginggpt_2023}
Yongliang Shen, Kaitao Song, Xu~Tan, Dongsheng Li, Weiming Lu, and Yueting
  Zhuang.
\newblock {HuggingGPT}: {Solving} {AI} {Tasks} with {ChatGPT} and its {Friends}
  in {Hugging} {Face}, May 2023.
\newblock arXiv:2303.17580 [cs].

\bibitem{shinn_reflexion_2023}
Noah Shinn, Federico Cassano, Beck Labash, Ashwin Gopinath, Karthik Narasimhan,
  and Shunyu Yao.
\newblock Reflexion: {Language} {Agents} with {Verbal} {Reinforcement}
  {Learning}, June 2023.
\newblock arXiv:2303.11366 [cs].

\bibitem{song_restgpt_2023}
Yifan Song, Weimin Xiong, Dawei Zhu, Cheng Li, Ke~Wang, Ye~Tian, and Sujian Li.
\newblock {RestGPT}: {Connecting} {Large} {Language} {Models} with
  {Real}-{World} {Applications} via {RESTful} {APIs}, June 2023.
\newblock arXiv:2306.06624 [cs].

\bibitem{taylor2022galactica}
Ross Taylor, Marcin Kardas, Guillem Cucurull, Thomas Scialom, Anthony
  Hartshorn, Elvis Saravia, Andrew Poulton, Viktor Kerkez, and Robert Stojnic.
\newblock Galactica: A large language model for science.
\newblock {\em arXiv preprint arXiv:2211.09085}, 2022.

\bibitem{valmeekam_planning_2023}
Karthik Valmeekam, Matthew Marquez, Sarath Sreedharan, and Subbarao
  Kambhampati.
\newblock On the {Planning} {Abilities} of {Large} {Language} {Models} -- {A}
  {Critical} {Investigation}, May 2023.
\newblock arXiv:2305.15771 [cs].

\bibitem{wang2023scientific}
Hanchen Wang, Tianfan Fu, Yuanqi Du, Wenhao Gao, Kexin Huang, Ziming Liu, Payal
  Chandak, Shengchao Liu, Peter Van~Katwyk, Andreea Deac, et~al.
\newblock Scientific discovery in the age of artificial intelligence.
\newblock {\em Nature}, 620(7972):47--60, 2023.

\bibitem{wang2023learning}
Qingyun Wang, Doug Downey, Heng Ji, and Tom Hope.
\newblock Learning to generate novel scientific directions with contextualized
  literature-based discovery.
\newblock {\em arXiv preprint arXiv:2305.14259}, 2023.

\bibitem{wang2022self}
Yizhong Wang, Yeganeh Kordi, Swaroop Mishra, Alisa Liu, Noah~A Smith, Daniel
  Khashabi, and Hannaneh Hajishirzi.
\newblock Self-instruct: Aligning language model with self generated
  instructions.
\newblock {\em arXiv preprint arXiv:2212.10560}, 2022.

\bibitem{wei_chain--thought_2023}
Jason Wei, Xuezhi Wang, Dale Schuurmans, Maarten Bosma, Brian Ichter, Fei Xia,
  Ed~Chi, Quoc Le, and Denny Zhou.
\newblock Chain-of-{Thought} {Prompting} {Elicits} {Reasoning} in {Large}
  {Language} {Models}, January 2023.
\newblock arXiv:2201.11903 [cs].

\bibitem{wei2022chain}
Jason Wei, Xuezhi Wang, Dale Schuurmans, Maarten Bosma, Fei Xia, Ed~Chi, Quoc~V
  Le, Denny Zhou, et~al.
\newblock Chain-of-thought prompting elicits reasoning in large language
  models.
\newblock {\em Advances in Neural Information Processing Systems},
  35:24824--24837, 2022.

\bibitem{wu2023pmcllama}
Chaoyi Wu, Weixiong Lin, Xiaoman Zhang, Ya~Zhang, Yanfeng Wang, and Weidi Xie.
\newblock Pmc-llama: Towards building open-source language models for medicine,
  2023.

\bibitem{xi_rise_2023}
Zhiheng Xi, Wenxiang Chen, Xin Guo, Wei He, Yiwen Ding, Boyang Hong, Ming
  Zhang, Junzhe Wang, Senjie Jin, Enyu Zhou, Rui Zheng, Xiaoran Fan, Xiao Wang,
  Limao Xiong, Yuhao Zhou, Weiran Wang, Changhao Jiang, Yicheng Zou, Xiangyang
  Liu, Zhangyue Yin, Shihan Dou, Rongxiang Weng, Wensen Cheng, Qi~Zhang,
  Wenjuan Qin, Yongyan Zheng, Xipeng Qiu, Xuanjing Huang, and Tao Gui.
\newblock The {Rise} and {Potential} of {Large} {Language} {Model} {Based}
  {Agents}: {A} {Survey}, September 2023.
\newblock arXiv:2309.07864 [cs].

\bibitem{yang_auto-gpt_2023}
Hui Yang, Sifu Yue, and Yunzhong He.
\newblock Auto-{GPT} for {Online} {Decision} {Making}: {Benchmarks} and
  {Additional} {Opinions}, June 2023.
\newblock arXiv:2306.02224.

\bibitem{yang2023large}
Zonglin Yang, Xinya Du, Junxian Li, Jie Zheng, Soujanya Poria, and Erik
  Cambria.
\newblock Large language models for automated open-domain scientific hypotheses
  discovery.
\newblock {\em arXiv preprint arXiv:2309.02726}, 2023.

\bibitem{yao_tree_2023}
Shunyu Yao, Dian Yu, Jeffrey Zhao, Izhak Shafran, Thomas~L. Griffiths, Yuan
  Cao, and Karthik Narasimhan.
\newblock Tree of {Thoughts}: {Deliberate} {Problem} {Solving} with {Large}
  {Language} {Models}, May 2023.
\newblock arXiv:2305.10601 [cs].

\bibitem{yao_react_2023}
Shunyu Yao, Jeffrey Zhao, Dian Yu, Nan Du, Izhak Shafran, Karthik Narasimhan,
  and Yuan Cao.
\newblock {ReAct}: {Synergizing} {Reasoning} and {Acting} in {Language}
  {Models}, March 2023.
\newblock arXiv:2210.03629.

\bibitem{zhang_sirens_2023}
Yue Zhang, Yafu Li, Leyang Cui, Deng Cai, Lemao Liu, Tingchen Fu, Xinting
  Huang, Enbo Zhao, Yu~Zhang, Yulong Chen, Longyue Wang, Anh~Tuan Luu, Wei Bi,
  Freda Shi, and Shuming Shi.
\newblock Siren's {Song} in the {AI} {Ocean}: {A} {Survey} on {Hallucination}
  in {Large} {Language} {Models}, September 2023.
\newblock arXiv:2309.01219 [cs].

\bibitem{zhong2023goal}
Ruiqi Zhong, Peter Zhang, Steve Li, Jinwoo Ahn, Dan Klein, and Jacob
  Steinhardt.
\newblock Goal driven discovery of distributional differences via language
  descriptions.
\newblock {\em arXiv preprint arXiv:2302.14233}, 2023.

\bibitem{zhu_ghost_2023}
Xizhou Zhu, Yuntao Chen, Hao Tian, Chenxin Tao, Weijie Su, Chenyu Yang, Gao
  Huang, Bin Li, Lewei Lu, Xiaogang Wang, Yu~Qiao, Zhaoxiang Zhang, and Jifeng
  Dai.
\newblock Ghost in the {Minecraft}: {Generally} {Capable} {Agents} for
  {Open}-{World} {Environments} via {Large} {Language} {Models} with
  {Text}-based {Knowledge} and {Memory}, June 2023.
\newblock arXiv:2305.17144 [cs].

\bibitem{zhu2023pad}
Xuekai Zhu, Biqing Qi, Kaiyan Zhang, Xingwei Long, and Bowen Zhou.
\newblock Pad: Program-aided distillation specializes large models in
  reasoning.
\newblock {\em arXiv preprint arXiv:2305.13888}, 2023.

\end{thebibliography}

\clearpage
\appendix
\section{Related Works} 

\subsection{Data-Driven Scientific Discovery}

Data-driven knowledge discovery research within LLM is relatively limited, with the current focus primarily on dataset construction and task-driven design.
In this context, \cite{zhong2023goal} proposed a dataset for investigating the transition from goals to discoveries. However, it should be noted that accurate discoveries within this dataset are not recent.
\cite{wang2023learning} introduced a method for automatically collecting and constructing publication data, along with a proposal for a hypothesis generation approach in the natural language processing (NLP) domain. However, this method requires prior human knowledge, explicit context, and is not an automated process. It's worth noting that their data was constructed from literature before 2021 from the ACL collection, implying that the information may already exist in open-source models like chatGPT and LLAMA.
Furthermore, \cite{wang2023learning} focused on integrating computational tools in the field of chemistry, primarily analyzing the capabilities of LLMs in using integrated tools but neglecting the ability for zero-shot generalization in chemistry reactions.
\cite{boiko2023emergent} delved more into the abilities of LLMs regarding planning and conducting experiments but did not consider proposing new hypotheses.
\cite{yang2023large} introduced a new task for open-domain hypothesis induction and created a dataset comprising 50 articles from social science journals. Additionally, they developed a multi-module system for exploring feedback mechanisms. 
However, all of the above-mentioned literature lacks strict guarantees on the visibility of test data to models, thereby limiting our exploration of the zero-shot generalization capability of LLMs through learning from existing knowledge to propose new hypothesis.
Unlike existing works, we have designed datasets based on publication dates, which can easily ensure a strict independence between test data and LLMs.

\subsection{LLM-driven Autonomous Agents}

Large language models demonstrate exceptional capabilities in tasks such as question answering, program coding, and instruction following. However, they still confront significant challenges related to factual hallucination~\cite{zhang_sirens_2023, rawte_survey_2023}, knowledge outdated~\cite{cao_life_2023}, and interactions with real-world.
To address these challenges, recent research has explored enhancing LLMs by incorporating tools such as search engines~\cite{nakano_webgpt_2022, parisi_talm_2022}, calculators~\cite{schick_toolformer_2023}, code interpreter~\cite{zhu2023pad}, RESTful APIs~\cite{song_restgpt_2023, patil_gorilla_2023} and others.
The integration of LLMs with tool use, also known as LLM-driven autonomous agents (LAAs), has attracted substantial public attention.
These agents are equipped with reasoning~\cite{wei_chain--thought_2023, yao_tree_2023}, planning~\cite{shen_hugginggpt_2023, valmeekam_planning_2023}, decision-making~\cite{yang_auto-gpt_2023, kang_think_2023, chen_asking_2023}, and long-term memory capabilities~\cite{zhu_ghost_2023, hu_chatdb_2023}, and they are constructed upon the foundation of LLMs.
LAAs can autonomously plan sub-goals for complex tasks, execute actions, obtain feedback from the environment, and adjust their behaviors to adapt~\cite{yao_react_2023, xi_rise_2023, shinn_reflexion_2023}.
LAAs have demonstrated significant potential in addressing complex real-world tasks, including software development~\cite{qian_communicative_2023, hong_metagpt_2023}, drama creation~\cite{maas_show-1_nodate}, course design~\cite{chen_empowering_2023}, chemistry experiments~\cite{bran_chemcrow_2023} and more.	
Furthermore, multi-agent collaboration plays a significant role in LAA applications, allowing agents to collaborate and interact to solve problems through various role-playing scenarios~\cite{park_generative_2023, fu_improving_2023, gong_mindagent_2023, li_camel_2023}.
To the best of our knowledge, there is still a dearth of exploration regarding the use of agents, particularly multi-agents, for scientific discovery.	
In this paper, our objective is to undertake a preliminary effort to enhance the hypothesis proposing capability of LLMs by harnessing tools and multiple agents, along with conducting an analysis of influencing factors.
\section{Implementation Details}
\label{apx:implementation_details}
In this section, we delve into further implementation details of our experiments, including information about the constructed dataset and open-source models.

\subsection{Details of Dataset}
\label{apx:details_dataset}

\begin{figure}[ht]
    \centering
    \begin{subfigure}[b]{0.35\textwidth}
        \centering
        \includegraphics[width=1.0\linewidth]{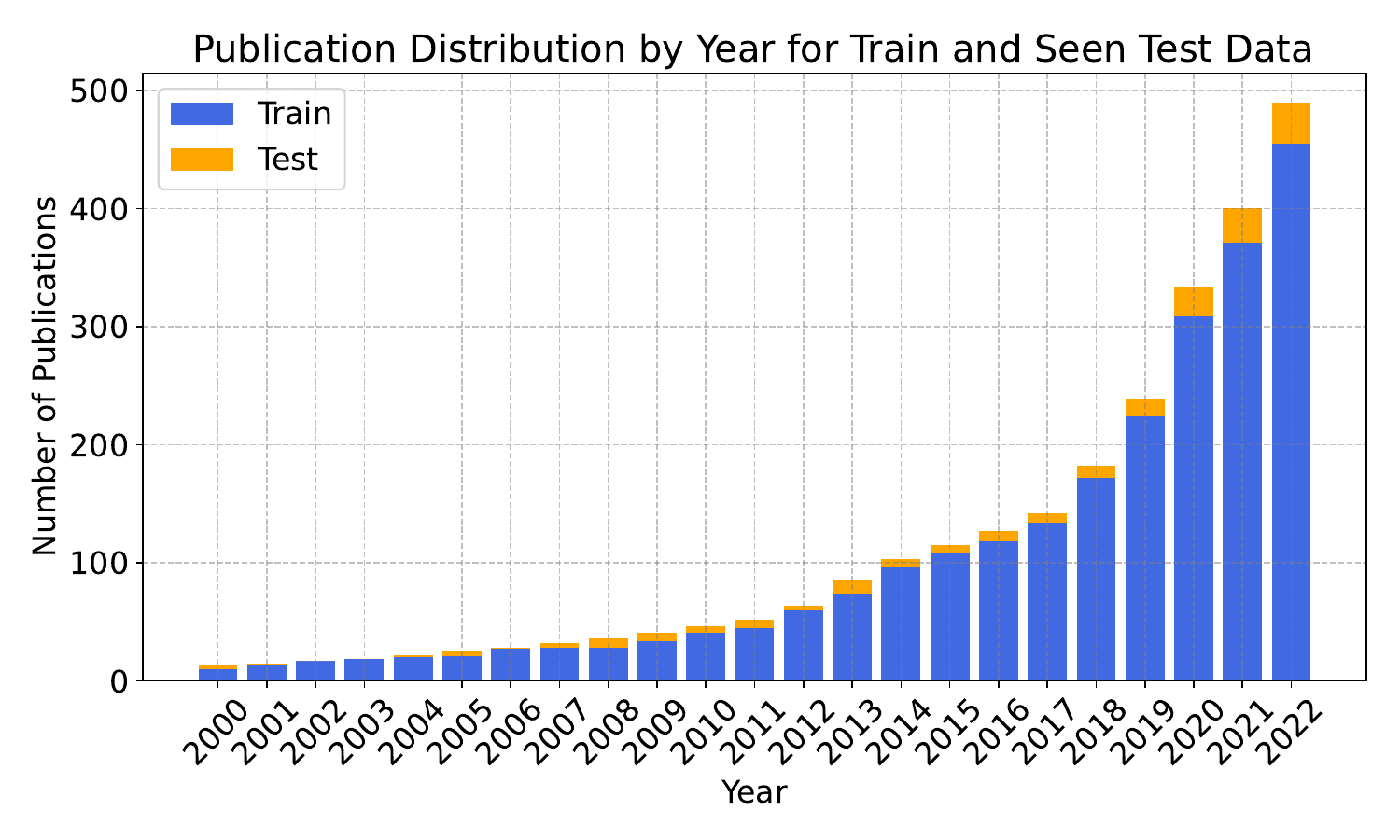}
        \label{fig:data_train}
    \end{subfigure}\hfill
    \begin{subfigure}[b]{0.35\textwidth}
        \centering
        \includegraphics[width=1.0\linewidth]{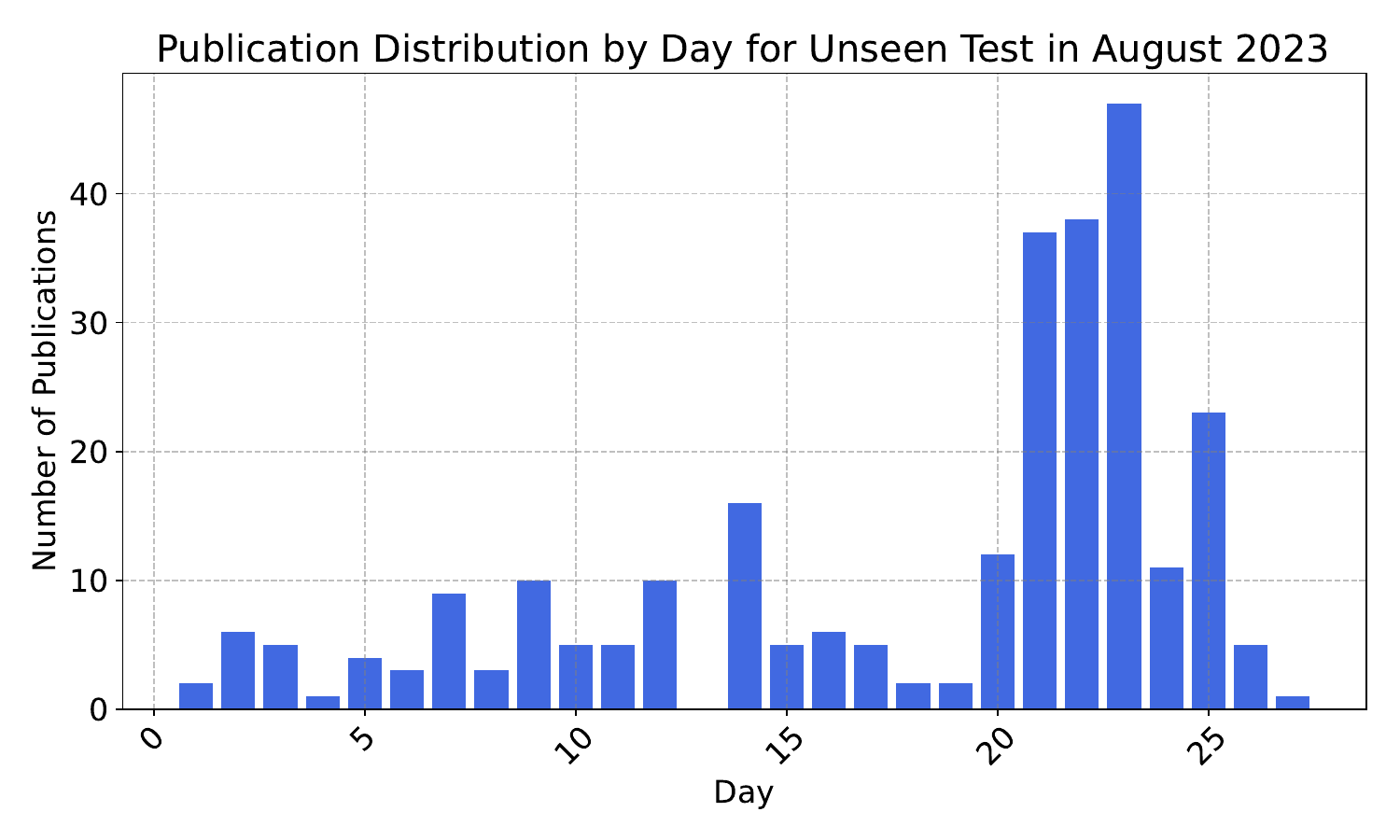}
        \label{fig:data_unseen}
    \end{subfigure}\hfill
    \begin{subfigure}[b]{0.25\textwidth}
        \centering
        \includegraphics[width=1.0\linewidth]{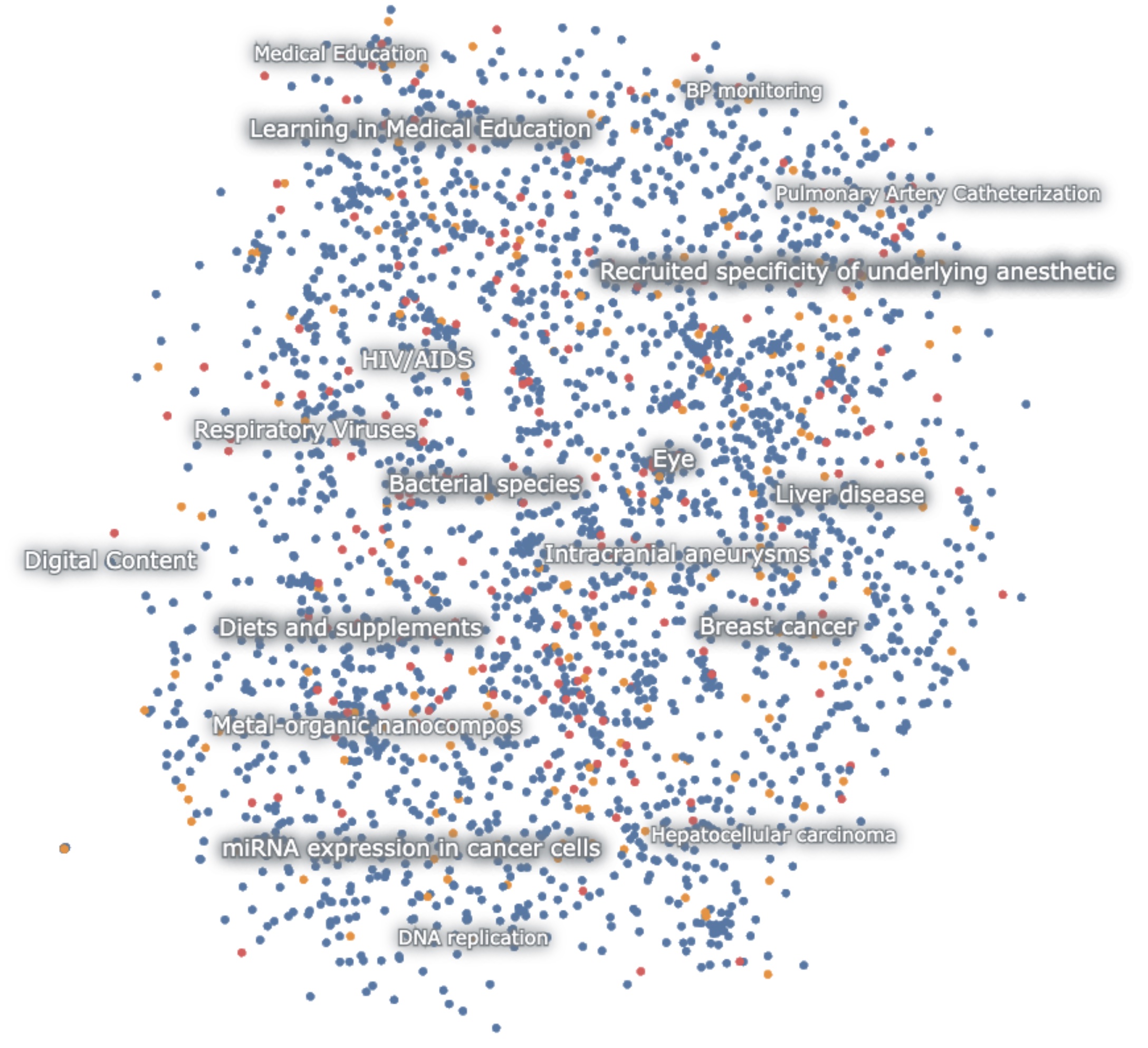}
        \label{fig:data_keywords}
    \end{subfigure}
    \caption{Distribution of the background and hypothesis pairs (BHP) dataset: In the left panel, we present the publication distribution by year for the training and seen test datasets, indicating a steady increase year by year until January 2023. In the center panel, we depict the publication distribution by month for the unseen test dataset, which was sampled from August 2023 and emphasizes the latter part of the month. The right panel displays the distribution of keywords in abstracts from the training, seen test, and unseen test datasets, represented by blue, yellow, and red, respectively.	}
    \label{fig:data_distribution}
\end{figure}

\textbf{Distribution of Training and Test Sets.} We present the publication dates and topic distributions of the various datasets for comparison, as illustrated in Figure~\ref{fig:data_distribution}, where we utilize \verb|Nomic Atlas|~\footnote{\url{https://github.com/nomic-ai/nomic}} to visualize the topic distribution of abstracts in both the training and test datasets.

\subsection{Details of Models}
\label{apx:details_models}
We present the meta-information of the open-source models used in our experiments, as shown in Table~\ref{tab:model_metadata}. We have gathered data regarding their pre-training, supervised learning corpus, and release dates to ensure the non-visibility of the unseen test data.

\begin{table}[ht]
  \caption{To further ensure the non-visibility of the test data, we provide an overview of the related literature corpus within the training set of various LLMs, accompanied by their respective publication dates. The data marked with (*) is the data generated by people talking to ChatGPT. Our date marking is consistent with ChatGPT.}
  \label{tab:model_metadata}
  \centering
  \scalebox{0.8}{
  \begin{tabular}{cllll}
    \toprule
    \hline
    \textbf{Category} & \textbf{Model} & Base Model & SFT Data (Y/M) & Released \\
    \midrule
    \multirow{ 3}{*}{API-based} &  
                     gpt-3.5-turbo (0-shot) & GPT-3 &  Unknown  &  2022/12  \\
                   & gpt-3.5-turbo (5-shot) & GPT-3 &  Unknown  &  2022/12  \\
                   & gpt-4*           & GPT-4 &  Unknown  &  2023/06  \\ \midrule
    \multirow{ 7}{*}{General} 
         & \href{https://huggingface.co/lmsys/vicuna-33b-v1.3}{Vicuna-33b-v1.3}
                                & Llama-1  & \href{https://sharegpt.com/}{ShareGPT} (Unknown)  &  2023/06    \\ 
         & \href{https://huggingface.co/meta-llama/Llama-2-7b-hf}{Llama-2-7b-chat} & Llama-2  & Unknown   &  2023/07  \\
         & \href{https://huggingface.co/meta-llama/Llama-2-13b-chat-hf}{Llama-2-13b-chat}   & Llama 2  & Unknown   &  2023/07  \\
         & \href{https://huggingface.co/meta-llama/Llama-2-70b-chat-hf}{Llama-2-70b-chat}   & Llama 2  & Unknown   &  2023/07  \\
         & \href{https://huggingface.co/WizardLM/WizardLM-13B-V1.2}{WizardLM-13B-V1.2}  & Llama-2  & \href{https://huggingface.co/datasets/WizardLM/WizardLM_evol_instruct_V2_196k/tree/main}{Alpaca and ShareGPT} (2023/06)    &  2023/07    \\
         & \href{https://huggingface.co/WizardLM/WizardLM-70B-V1.0}{WizardLM-70B-V1.0}  & Llama-2  & \href{https://huggingface.co/datasets/WizardLM/WizardLM_evol_instruct_V2_196k/tree/main}{Alpaca and ShareGPT} (2023/06)    &  2023/08    \\
          & \href{https://huggingface.co/openchat/openchat_v3.2_super}{openchat-v3.2-super}  & Llama-2  &  \href{https://huggingface.co/datasets/openchat/openchat_sharegpt4_dataset/tree/main}{Sharegpt4 Dataset} (2023/06)    &  2023/09    \\ \midrule
    \multirow{ 3}{*}{Medicine} 
         & \href{https://github.com/kbressem/medAlpaca}{MedAlpaca-13B} & Llama-1* & Mixture (2023/03) & 2023/03 \\
         & \href{https://github.com/Kent0n-Li/ChatDoctor}{ChatDoctor*} & Llama-1* & Mixture (2023/04) & 2023/04 \\
         & \href{https://github.com/chaoyi-wu/PMC-LLaMA}{PMC-LLaMA-13B}& Llama-2* & Mixture (2023/04) & 2023/08* \\
    \hline
    \bottomrule
  \end{tabular}}
\end{table}

\section{Additional Results}
\label{apx:additional_results}

We have included additional results from human evaluations in Table~\ref{tab:additional_human_results}, primarily focusing on correlation scores between word overlap metrics and manual evaluations. 
Note that we continue to use the same samples used in human evaluation to compute BLEU and ROUGE-L for a fair comparison.
We calculate the Pearson and Spearman coefficients between each automatic metric and the average human score.
These results reveal that word overlap metrics, such as BLEU and ROUGE-L, exhibit notably lower correlation with manual scores. While BLEU and ROUGE-L may have a high correlation with relevance metrics, they are weak in providing a comprehensive evaluation of the generations.
Conversely, evaluations conducted by ChatGPT demonstrate higher correlation with human evaluations, as illustrated in Table~\ref{tab:human_eval}. 
However, there is still a significant need to explore advanced metrics, particularly automated ones, in the context of scientific discovery.

\begin{table}[htbp]
  \small
  \caption{
  The table illustrates the correlations between automatic metrics and human evaluations. We annotate the Pearson and Spearman scores after each correlation score, denoting them as $r$ and $\rho$.
  }
  \label{tab:additional_human_results}
  \centering
  \scalebox{0.85}{
  \begin{tabular}{clcccc}
    \toprule
    \hline
    \hfill
     \multirow{2}{*}{Category} & \multirow{2}{*}{Model} & \multicolumn{2}{c}{Word Overlap} & ChatGPT & Human\\
    \cmidrule(lr){3-4} \cmidrule(lr){5-5} \cmidrule(lr){6-6} 
      &   & BLEU ($r$/$\rho$) & ROUGE-L ($r$/$\rho$) &  Avg ($r$/$\rho$)  & Avg ($r$/$\rho$)  \\
    \midrule
     \multirow{2}{*}{API-based} 
     & gpt-3.5-turbo(0-shot)    & 16.59(0.03/0.01)  & 29.87(-0.04/-0.05) & 1.90(0.87/0.78)  &  2.02(1.00/1.00) \\
     & gpt-3.5-turbo(5-shot)*   & 14.99(-0.09/0.12) & 27.51(-0.33/-0.35) & 1.96(0.80/0.78)  &  2.15(1.00/1.00) \\
     \midrule
     \multirow{5}{*}{General} 
     & Llama-2-70b-chat(0-shot) & 9.64(-0.21/-0.20) & 22.17(-0.31/-0.28) & 2.04(0.89/0.84)  &  1.96(1.00/1.00) \\
     & Llama-2-70b-chat(5-shot) & 9.42(-0.58/-0.65) & 20.59(-0.47/-0.42) & 2.20(0.96/0.90)  &  2.33(1.00/1.00) \\
     & Llama-2-70b-chat(5-shot)*& 9.60(-0.16/-0.10) & 19.99(-0.15/-0.17) & 2.01(0.97/0.94)  &  2.08(1.00/1.00) \\
     & WizardLM-70B-V1.0(0-shot)& 11.42(0.21/0.36) & 24.11(0.29/0.49) & 1.91(0.90/0.75)  &  1.81(1.00/1.00) \\
     & WizardLM-70B-V1.0(5-shot)& 9.86(-0.28/-0.37) & 23.52(-0.17/-0.24) & 2.01(0.85/0.89)  &  2.02(1.00/1.00) \\
     \midrule
     \multirow{2}{*}{Medicine} 
     & PMC-LLaMA-13B(0-shot)    & 8.19(0.32/0.39) & 21.85(0.18/0.27) & 1.41(0.73/0.73)  &  1.88(1.00/1.00) \\
     & PMC-LLaMA-13B(5-shot)*   & 5.52(0.06/-0.01) & 13.64(0.26/0.23) & 1.97(0.95/0.94)  &  1.92(1.00/1.00) \\
     \midrule
     SFT & WizardLM-13B-V1.2     & 21.48(-0.00/0.00) & 27.83(0.17/0.27) & 1.79(0.83/0.85)  &  1.77(1.00/1.00) \\
    \hline
    \bottomrule
  \end{tabular}}
\end{table}

\section{Case Study}
\label{apx:case_study}
In this section, we present several generated hypotheses from various models and provide examples of the evaluation process, step by step, using ChatGPT.

\subsection{Generated Hypothesis}
We compare the generated hypotheses of different LLMs selected in human evaluation. The selected medicine literature was published in August 2023~\cite{changNewMechanicalMarkers2023}, which proposed the power law index as an early marker of myocardial infarction. As shown in Table~\ref{tab:appendix-case-llm}, some responses like \verb|gpt-3.5-turbo (0-shot)| propose this hypothesis from zero, based only on the background. This indicates that LLMs have great potential in proposing highly novel and verifiable hypotheses and could play a significant role in scientific discovery.
Furthermore, we showcase the proposed hypotheses in a multi-agent collaboration involving three roles: Analyst, Scientist, and Critic, working together to generate hypotheses across 6 rounds, as shown in Table~\ref{tab:appendix-case-multi-agent}.

\begin{longtable}{p{13.5cm}}
\caption{
This table illustrates the hypothesis proposals from different LLMs.
}
\label{tab:appendix-case-llm}
\endfirsthead
\endhead
\toprule
\hline
\small
\underline{\textbf{\textsc{Instructed LLMs Case}}} \\
\textbf{Background:}\\
(1) In the early stage of MI, $\alpha$short and $\alpha$long can track and distinguish MI samples from healthy control samples with an AUC > 0.7. \\
(2) In the mid-stage and late-stage of MI, E2, E3, and Etotal exhibit excellent discriminating performance with an AUC > 0.6.\\ 
(3) All cardiac tissues display a universal two-stage power-law rheology at different time scales, independent of their disease status. \\
\\
\textbf{Golden Hypothesis:}\\
(1) The sequential order of MI-related changes in cardiac tissues starts from the cytoplasmic level and progresses to the cellular and tissue levels. \\
(2) Pathological changes at subcellular, cellular, and tissue levels are accompanied by mechanical alterations. \\
(3) The proposed mechanical markers have diagnostic potential for assessing the progression of MI and studying other cardiovascular diseases. \\
\\
\hline
\\
\textbf{gpt-3.5-turbo (0-shot):}\\
(1) The combination of $\alpha$short and $\alpha$long can be used as a reliable biomarker for early detection and differentiation of myocardial infarction (MI) from healthy controls. \\
(2) E2, E3, and Etotal can serve as effective biomarkers for distinguishing between mid-stage and late-stage MI patients with high accuracy. \\
(3) The universal two-stage power-law rheology observed in cardiac tissues may play a crucial role in maintaining tissue integrity and function, regardless of the presence or absence of disease.\\
\\
\hline
\\
\textbf{gpt-3.5-turbo (5-shot):}\\
(1) The levels of $\alpha$short and $\alpha$long may serve as potential biomarkers for early detection and diagnosis of myocardial infarction (MI). \\
(2) E2, E3, and Etotal may have utility as biomarkers for assessing the progression and severity of MI in the mid and late stages. \\
(3) The universal two-stage power-law rheology observed in cardiac tissues suggests a fundamental biomechanical principle underlying tissue behavior, regardless of disease status.\\
\\
\hline
\\
\textbf{Llama-2-70b-chat (0-shot):}\\
(1) E2, E3, and Etotal can also track and distinguish MI samples from healthy control samples in the early stage of MI, with an AUC > 0.7, similar to $\alpha$short and $\alpha$long.\\
(2) The universal two-stage power-law rheology of cardiac tissues is related to the alteration of the extracellular matrix and the degradation of collagen in the early stage of MI, which can be detected by E2, E3, and Etotal.\\
(3) The discriminating performance of E2, E3, and Etotal in the mid-stage and late-stage of MI is affected by the changes in the levels of collagen and proteoglycans in the extracellular matrix, which can be evaluated by measuring the mechanical properties of cardiac tissues using a uniaxial tensiometer.\\
\\
\hline
\\
\textbf{Llama-2-70b-chat (5-shot):}\\
(1) The discriminating performance of $\alpha$short and $\alpha$long in the early stage of MI is attributed to their ability to capture the subtle changes in cardiac tissue mechanics caused by the onset of fibrosis.\\
(2) E2, E3, and Etotal exhibit excellent discriminating performance in the mid-stage and late-stage of MI due to their sensitivity to changes in collagen content and cross-linking, which are hallmarks of fibrosis progression.\\
(3) The universal two-stage power-law rheology observed in all cardiac tissues, independent of disease status, suggests a common mechanism of mechanical regulation that can be exploited for early disease detection and monitoring. \\
\\
\hline
\\
\textbf{Llama-2-70b-chat (5-shot)*:}\\
(1) The $\alpha$short and $\alpha$long parameters can detect changes in myocardial stiffness in the early stages of myocardial infarction (MI). \\
(2) E2, E3, and Etotal can differentiate between healthy controls and MI samples in the mid and late stages. \\
(3) The universal two-stage power-law rheology in cardiac tissues may be useful in developing novel therapeutic approaches for MI.\\
\\
\hline
\\
\textbf{WizardLM-70B-V1.0 (0-shot):}\\
(1) $\alpha$short and $\alpha$long can track and distinguish MI samples from healthy control samples in all stages of MI with an AUC > 0.7.\\
(2) E2, E3, and Etotal exhibit excellent discriminating performance in the early stage of MI with an AUC > 0.6.\\
(3) The two-stage power-law rheology observed in all cardiac tissues is related to the progression of MI.\\
\\
\hline
\\
\textbf{WizardLM-70B-V1.0 (5-shot):}\\
(1) The discriminating performance of $\alpha$short and $\alpha$long in the early stage of MI may be attributed to their sensitivity to early molecular changes in the myocardium. \\
(2) The discriminating performance of E2, E3, and Etotal in the mid-stage and late-stage of MI may be due to their ability to reflect the progression of myocardial damage and remodeling. \\
(3) The universal two-stage power-law rheology observed in all cardiac tissues may be a fundamental property of cardiac tissue biomechanics, potentially useful for developing diagnostic and therapeutic strategies for MI.\\
\\
\hline
\\
\textbf{PMC-LLaMA-13B (0-shot):}\\
(1) $\alpha$short and $\alpha$long are effective biomarkers for distinguishing MI samples from healthy control samples both in the early stage and the late stage of MI. \\
(2) E2 and E3 (Etotal) are likely to serve as effective biomarkers for discriminating mid-stage or late-stage MI samples from healthy control samples. \\
(3) Cardiac tissues display a universal two-stage power-law rheology that is independent of their disease status.\\
\\
\hline
\\
\textbf{PMC-LLaMA-13B (5-shot):}\\
(1) Further comprehensive validation of the proposed framework is needed to elucidate the pathological changes in MI patients.\\
(2) Consideration of multiple markers is important. \\
(3) The influence of various external factors should be incorporated for accurate interpretation of the mechanical markers.\\
\\
\hline
\\
\textbf{WizardLM-13B-V1.2 (SFT):} \\
(1) The combination of $\alpha$short and $\alpha$long features can provide a more robust identification of MI samples.\\
(2) E2, E3, and Etotal features can serve as independent predictors of MI.\\
(3) The universal two-stage power-law rheology observed in all cardiac tissues may have diagnostic implications.\\
    \hline
    \bottomrule
\end{longtable}

\begin{longtable}{p{13.5cm}}
\caption{
This table illustrates a case from multi-agent collaboration on hypothesis proposing.
}
\label{tab:appendix-case-multi-agent}
\endfirsthead
\endhead
\toprule
\hline
\small
\underline{\textbf{\textsc{Multi-Agent Case}}} \\
\textbf{Background:}\\
 (1) The past 25 years have seen significant development in the field of digital electrophysiology, including advancements in telemonitoring of cardiac implantable electronic devices (CIEDs) and the use of wearable devices for cardiac rhythm monitoring.\\
 (2) Telemonitoring of CIEDs has been shown to reduce time to detection of clinically actionable events and improve patient outcomes. \\
 (3)Wearable devices, such as smartwatches, rings, and fitness trackers, have become more accurate and reliable in monitoring cardiac rhythm using photoplethysmography.  
\\
\\
\textbf{Golden Hypothesis:}\\
 (1) The implementation of telemonitoring and the use of wearable devices for cardiac rhythm monitoring will continue to increase in the future. \\
 (2) Artificial intelligence can be utilized to better identify actionable alerts and improve patient outcomes in telemonitoring of CIEDs. \\
 (3) Smartwatches may have the potential to replace CIEDs in certain cases.  
\\
\\
\textbf{Generated Hypothesis:}\\
(1) The advancements in telemonitoring of cardiac implantable electronic devices (CIEDs) and the use of wearable devices for cardiac rhythm monitoring have led to improved patient outcomes, such as reduced hospitalizations, improved quality of life, and decreased mortality rates.\\
(2) These advancements achieve these outcomes by enabling early detection of abnormal rhythms, facilitating timely interventions, and providing valuable data for personalized treatment plans.\\
(3) This hypothesis is supported by studies that have demonstrated the effectiveness of telemonitoring in reducing time to detection of clinically actionable events and improving patient outcomes.The refined hypothesis captures the essence of the research background and the collaborative insights, providing a logical and scientifically grounded statement. It can now be further evaluated and tested in future research. \\
\\
\hline
\\
\underline{\textbf{Dialogue process of multi-agent collaboration.}}\\
\textbf{\textit{(Round 1) Analyst:}} Based on the provided research background, I have identified the following keywords and topics:\\
\\
Keywords:\\
- Digital electrophysiology\\
- Telemonitoring\\
- Cardiac implantable electronic devices (CIEDs)\\
- Wearable devices\\
- Cardiac rhythm monitoring\\
- Photoplethysmography\\
\\
Topics:\\
1. Advancements in digital electrophysiology and telemonitoring of CIEDs\\
2. Benefits of telemonitoring in improving patient outcomes\\
3. Accuracy and reliability of wearable devices for cardiac rhythm monitoring using photoplethysmography\\
\\
Preliminary insights:\\
- The field of digital electrophysiology has seen significant development in the past 25 years, particularly in the areas of telemonitoring of CIEDs and the use of wearable devices for cardiac rhythm monitoring.\\
- Telemonitoring of CIEDs has been shown to have positive impacts, such as reducing the time to detect clinically actionable events and improving patient outcomes.\\
- Wearable devices, including smartwatches, rings, and fitness trackers, have become more accurate and reliable in monitoring cardiac rhythm using photoplethysmography.\\
\\
Based on these insights, it can be hypothesized that the advancements in telemonitoring and wearable devices have led to improved cardiac rhythm monitoring and patient outcomes. However, further analysis and collaboration with the Scientist and Critic are needed to refine and validate this hypothesis.\\
\\
\hline
\\
\textbf{\textit{(Round 2) Scientist:}} Based on the insights provided by the Analyst and the research background, I propose the following hypothesis:\\
\\
Hypothesis: The advancements in telemonitoring of cardiac implantable electronic devices (CIEDs) and the use of wearable devices for cardiac rhythm monitoring have led to improved patient outcomes and more accurate detection of clinically actionable events.\\
\\
This hypothesis suggests that the development and implementation of telemonitoring technologies for CIEDs, along with the increasing accuracy and reliability of wearable devices using photoplethysmography, have positively impacted patient outcomes by reducing the time to detect clinically actionable events. By continuously monitoring cardiac rhythm, these advancements have the potential to provide timely interventions and improve overall patient care.\\
\\
The hypothesis is grounded in the existing literature, which highlights the benefits of telemonitoring in improving patient outcomes and the increasing accuracy of wearable devices for cardiac rhythm monitoring. However, further collaboration with the Critic is necessary to evaluate the novelty, coherence, and scientific validity of the hypothesis.\\
\\
\hline
\\
\textbf{\textit{(Round 3) Critic:}} The hypothesis proposed by the Scientist is well-aligned with the research background and the insights provided by the Analyst. It effectively captures the essence of the advancements in telemonitoring of CIEDs and the use of wearable devices for cardiac rhythm monitoring. The hypothesis also highlights the potential positive impact on patient outcomes and the detection of clinically actionable events.\\
\\
However, there are a few areas that could be further refined or expanded upon in the hypothesis:\\
\\
1. **Specific outcomes**: The hypothesis mentions ""improved patient outcomes,"" but it would be beneficial to specify the specific outcomes that are expected to improve. For example, are we referring to reduced hospitalizations, improved quality of life, or decreased mortality rates?\\
\\
2. **Mechanisms**: The hypothesis briefly mentions the potential for timely interventions and improved patient care, but it would be helpful to elaborate on the specific mechanisms through which the advancements in telemonitoring and wearable devices are expected to achieve these outcomes. Are there specific interventions or treatments that can be administered based on the data collected from these devices?\\
\\
3. **Scientific validity**: While the hypothesis is grounded in the existing literature, it would be valuable to provide some references or studies that support the claims made in the hypothesis. This would enhance the scientific validity of the hypothesis and demonstrate that it is not just speculative but based on empirical evidence.\\
\\
To refine the hypothesis, I would recommend the Scientist to consider incorporating these suggestions and providing more specific details and references to support the claims made. This will strengthen the coherence and scientific validity of the hypothesis.\\
\\
Analyst, based on the feedback provided, please re-evaluate the research background and provide any additional insights or interpretations that could further refine the hypothesis.\\
\\
\hline
\\
\textbf{\textit{(Round 4) Analyst:}} Based on the feedback from the Critic, I have re-evaluated the research background and identified additional insights:\\
\\
1. **Specific outcomes**: The research background mentions that telemonitoring of CIEDs has been shown to reduce time to detection of clinically actionable events and improve patient outcomes. These outcomes could include early detection of arrhythmias, reduced hospitalizations due to cardiac events, improved quality of life, and decreased mortality rates. It would be beneficial to specify one or more of these specific outcomes in the hypothesis.\\
\\
2. **Mechanisms**: The advancements in telemonitoring and wearable devices can potentially lead to improved patient outcomes through several mechanisms. For example, continuous monitoring of cardiac rhythm can enable early detection of abnormal rhythms, allowing for timely interventions such as medication adjustments or electrical cardioversion. Additionally, the data collected from these devices can provide valuable insights for personalized treatment plans and remote patient management. Including these specific mechanisms in the hypothesis would enhance its coherence.\\
\\
3. **Scientific validity**: To strengthen the scientific validity of the hypothesis, it would be valuable to reference specific studies or research that support the claims made. For example, citing studies that have demonstrated the effectiveness of telemonitoring in reducing time to detection of clinically actionable events or improving patient outcomes would provide empirical evidence to support the hypothesis.\\
\\
Incorporating these additional insights and suggestions, the refined hypothesis could be:\\
\\
Hypothesis: The advancements in telemonitoring of cardiac implantable electronic devices (CIEDs) and the use of wearable devices for cardiac rhythm monitoring have led to improved patient outcomes, such as reduced hospitalizations, improved quality of life, and decreased mortality rates. These advancements achieve these outcomes by enabling early detection of abnormal rhythms, facilitating timely interventions, and providing valuable data for personalized treatment plans. This hypothesis is supported by studies that have demonstrated the effectiveness of telemonitoring in reducing time to detection of clinically actionable events and improving patient outcomes.\\
\\
This refined hypothesis takes into account the specific outcomes, mechanisms, and scientific validity, as suggested by the Critic. It provides a more comprehensive and well-supported statement that captures the essence of the research background and the collaborative insights.\\
\\
\hline
\\
\textbf{\textit{(Round 5) Scientist:}} Based on the additional insights and suggestions provided by the Analyst and the Critic, I propose the following refined hypothesis:\\
\\
Hypothesis: The advancements in telemonitoring of cardiac implantable electronic devices (CIEDs) and the use of wearable devices for cardiac rhythm monitoring have led to improved patient outcomes, such as reduced hospitalizations, improved quality of life, and decreased mortality rates. These advancements achieve these outcomes by enabling early detection of abnormal rhythms, facilitating timely interventions, and providing valuable data for personalized treatment plans. This hypothesis is supported by studies that have demonstrated the effectiveness of telemonitoring in reducing time to detection of clinically actionable events and improving patient outcomes.\\
\\
This refined hypothesis incorporates the specific outcomes mentioned in the research background, such as reduced hospitalizations, improved quality of life, and decreased mortality rates. It also expands on the mechanisms through which these advancements are expected to achieve these outcomes, including early detection of abnormal rhythms, timely interventions, and personalized treatment plans. Additionally, the hypothesis acknowledges the importance of scientific validity by referencing studies that support the claims made.\\
\\
The refined hypothesis is both innovative and scientifically grounded, as it builds upon the existing literature and incorporates the collaborative insights provided by the Analyst and the Critic. It provides a logical and well-supported statement that can be further evaluated and tested.\\
\\
\hline
\\
\textbf{\textit{(Round 6) Critic:}} The refined hypothesis proposed by the Scientist is well-structured and addresses the feedback provided by the Critic. It incorporates specific outcomes, mechanisms, and scientific validity, making it a comprehensive and scientifically robust statement.\\
\\
Final Answer:\\
(1) The advancements in telemonitoring of cardiac implantable electronic devices (CIEDs) and the use of wearable devices for cardiac rhythm monitoring have led to improved patient outcomes, such as reduced hospitalizations, improved quality of life, and decreased mortality rates.\\
(2) These advancements achieve these outcomes by enabling early detection of abnormal rhythms, facilitating timely interventions, and providing valuable data for personalized treatment plans.\\
(3) This hypothesis is supported by studies that have demonstrated the effectiveness of telemonitoring in reducing time to detection of clinically actionable events and improving patient outcomes.\\
\\
The refined hypothesis captures the essence of the research background and the collaborative insights, providing a logical and scientifically grounded statement. It can now be further evaluated and tested in future research.\\
    \hline
    \bottomrule
\end{longtable}

\subsection{Evaluation Examples}
We provide evaluation results conducted by ChatGPT in Table~\ref{tab:appendix-case-chatgpt-eval}. 
ChatGPT offers a step-by-step explanation for each metric, ultimately providing independent scores in integer format.
\begin{longtable}{p{13.5cm}}
\caption{
    This table illustrates a case for hypothesis evaluation using ChatGPT.
}
\label{tab:appendix-case-chatgpt-eval}
\endfirsthead
\endhead
\toprule
\hline
\small\
\underline{\textbf{\textsc{Case of ChatGPT as Annotator}}} \\
\textbf{Background:}\\
 (1) The past 25 years have seen significant development in the field of digital electrophysiology, including advancements in telemonitoring of cardiac implantable electronic devices (CIEDs) and the use of wearable devices for cardiac rhythm monitoring. \\
 (2) Telemonitoring of CIEDs has been shown to reduce time to detection of clinically actionable events and improve patient outcomes. \\
 (3)Wearable devices, such as smartwatches, rings, and fitness trackers, have become more accurate and reliable in monitoring cardiac rhythm using photoplethysmography. \\
 \\
 \textbf{Golden Hypothesis:}\\
  (1) The implementation of telemonitoring and the use of wearable devices for cardiac rhythm monitoring will continue to increase in the future. \\
  (2) Artificial intelligence can be utilized to better identify actionable alerts and improve patient outcomes in telemonitoring of CIEDs. \\
  (3) Smartwatches may have the potential to replace CIEDs in certain cases.  \\
\\
\textbf{Generated Hypothesis:}\\
(1) The advancements in telemonitoring of cardiac implantable electronic devices (CIEDs) and the use of wearable devices for cardiac rhythm monitoring have led to improved patient outcomes, such as reduced hospitalizations, improved quality of life, and decreased mortality rates.\\
(2) These advancements achieve these outcomes by enabling early detection of abnormal rhythms, facilitating timely interventions, and providing valuable data for personalized treatment plans.\\
(3) This hypothesis is supported by studies that have demonstrated the effectiveness of telemonitoring in reducing time to detection of clinically actionable events and improving patient outcomes.The refined hypothesis captures the essence of the research background and the collaborative insights, providing a logical and scientifically grounded statement. It can now be further evaluated and tested in future research.\\
\\
\midrule
\\
\underline{\textbf{Evaluation Results}} \\
\textbf{Novelty:} \\
Explanation: The generated scientific hypothesis builds upon the background information by stating that the advancements in telemonitoring of CIEDs and the use of wearable devices for cardiac rhythm monitoring have led to improved patient outcomes. It further suggests that these advancements achieve these outcomes by enabling early detection of abnormal rhythms, facilitating timely interventions, and providing valuable data for personalized treatment plans. This hypothesis goes beyond the background by proposing specific mechanisms through which the advancements in telemonitoring and wearable devices contribute to improved patient outcomes. It also mentions the supporting evidence from studies that have demonstrated the effectiveness of telemonitoring in reducing time to detection of clinically actionable events and improving patient outcomes. Overall, the hypothesis provides new insights and expands upon the existing knowledge in the field.\\
Score: 2\\
\\
\midrule
\\
\textbf{Relevance:} \\
Explanation: The generated scientific hypothesis is strongly related to the given background. The background provides information about the advancements in telemonitoring of cardiac implantable electronic devices (CIEDs) and the use of wearable devices for cardiac rhythm monitoring. The hypothesis builds upon this information by stating that these advancements have led to improved patient outcomes, such as reduced hospitalizations, improved quality of life, and decreased mortality rates. It further explains that these outcomes are achieved by enabling early detection of abnormal rhythms, facilitating timely interventions, and providing valuable data for personalized treatment plans. The hypothesis is supported by the background information, which mentions that telemonitoring of CIEDs has been shown to reduce time to detection of clinically actionable events and improve patient outcomes. Overall, the hypothesis is directly related to the background and provides a logical and scientifically grounded statement.\\
Score: 3\\
\\
\midrule
\\
\textbf{Significance:} \\
Explanation: The generated scientific hypothesis builds upon the background information provided and presents a logical and scientifically grounded statement. It suggests that the advancements in telemonitoring of CIEDs and the use of wearable devices for cardiac rhythm monitoring have led to improved patient outcomes, such as reduced hospitalizations, improved quality of life, and decreased mortality rates. The hypothesis further suggests that these advancements achieve these outcomes by enabling early detection of abnormal rhythms, facilitating timely interventions, and providing valuable data for personalized treatment plans. This hypothesis is supported by the background information that mentions the effectiveness of telemonitoring in reducing time to detection of clinically actionable events and improving patient outcomes. Overall, the hypothesis provides significant insights beyond the background and presents a testable statement for future research. \\
Score: 3 \\
\\
\midrule
\\
\textbf{Verifiability:} \\
Explanation: The generated scientific hypothesis is highly verifiable based on the given background. The background provides evidence of significant advancements in telemonitoring of cardiac implantable electronic devices (CIEDs) and the use of wearable devices for cardiac rhythm monitoring. It also states that telemonitoring of CIEDs has been shown to reduce time to detection of clinically actionable events and improve patient outcomes. Additionally, it mentions that wearable devices have become more accurate and reliable in monitoring cardiac rhythm. \\
\\
The hypothesis builds upon this background by suggesting that these advancements have led to improved patient outcomes, such as reduced hospitalizations, improved quality of life, and decreased mortality rates. It further proposes that these outcomes are achieved by enabling early detection of abnormal rhythms, facilitating timely interventions, and providing valuable data for personalized treatment plans. \\
\\
The hypothesis is supported by studies that have demonstrated the effectiveness of telemonitoring in reducing time to detection of clinically actionable events and improving patient outcomes. Therefore, the hypothesis is highly verifiable and has a strong likelihood of being confirmed in future research.\\
Score: 3 \\ 
    \hline
    \bottomrule
\end{longtable}
\section{Prompts Design}
\label{apx:prompt_design}
In this section, we offer prompts for prompt LLMs for hypothesis generation and for ChatGPT in the evaluation process.

\subsection{Zero- and Few-shot Prompts}
We provide prompts for hypothesis generation under both zero-shot and few-shot settings, which are detailed in Table~\ref{tab:appendix-context-1} and Table~\ref{tab:appendix-context-2} respectively.
Specially, the latter includes two methods to obtain in-context examples: random sampling and similarity retrieval.

\begin{table*}[h]
\caption{
Prompt for zero-shot hypothesis generation.
}
\label{tab:appendix-context-1}
\centering
\small
\begin{tabular}{p{\linewidth}}
\toprule
\hline
\underline{\textbf{\textsc{Zero-shot Instruction for hypothesis generation.}}} \\
    You are a researcher. You can give novel hypothesis based on your exist knowledge and the given background.\\
    Based on the known existing knowledge, generate new conjectures in the following format:\\
\\
(1) xxx\\
(2) xxx\\
(3) xxx\\
\\
Be sure to use English answers (proper nouns need to be marked in English), statements as concise and academic as possible, do not have too much repetitive information, numerical values using the original numbers, be sure to strictly follow the format, the corresponding content output to xxx.\\
\\
Note: Please respond directly to the multiple hypotheses without adding any extra sentences.
\\
Now give hypothesis based on the following background: \\
\{user\_input\}\\
        \hline
        \bottomrule
    \end{tabular}
\end{table*}

\begin{table*}
    \caption{
    Manually constructed context examples of background-hypothesis pairs sampling from literatures before January 2023.
    }
    \label{tab:appendix-context-2}
    \centering
    \small
\begin{tabular}{p{\linewidth}}
\toprule
\hline
\underline{\textbf{\textsc{Few-shot examples for hypothesis generation.}}} \\
    You are a renowned biomedical researcher. You can give novel hypothesis for the background based on your exist knowledge. Please follow the given examples and give the hypothesis in the SINGLE TURN. \\
\\
    Background:  \\
    (1) Neonatal intensive care is associated with long-term health problems in children such as cerebral palsy, mental retardation, deafness, blindness, learning disabilities, and behavioral problems. \\
    (2) Mothers of preterm infants experience more severe psychological distress compared to mothers of healthy full-term infants, but the impact of caregiving on parents of children discharged from NICUs is not well-researched. \\
    (3) Parents of NICU children show no difference in psychosocial health compared to parents of healthy full-term children.  \\
    Hypothesis: \\
    (1) The mental health of parents of NICU children may improve over time due to adaptation and relief from initial fear and anxiety.\\
    (2) Child characteristics, such as health status, behavior problems, and birth-related risk factors, may influence parental psychosocial health.\\
    (3) Certain factors, such as caregiver strain, family function, and demographic variables, may predict parental psychosocial health.  \\
\\
    Background:  \\
    (1) Recruitment of tumor supporting stromal cells and tissue remodeling in the tumor microenvironment support cancer cell proliferation, invasion, metastasis, and drug resistance. \\
    (2) Mesenchymal stem cells (MSC) are recruited by cancer cells into the tumor site and play a role in modulating tumor progression. \\
    (3) Intratumoral heterogeneity exists in solid tumors, with cancer stem cells (CSCs) and clonal evolution contributing to the complexity of cancer.  
    Hypothesis: \\
    (1) Transcriptional regulators are responsible for tumor-supporting stromal reprogramming, specifically in MSC in the tumor stroma. \\
    (2) Intercellular communication between cancer cells and recruited MSCs is mediated by cell-to-cell contact, paracrine interactions, and microvesicles. \\
    (3) Epithelial cancer cell plasticity is regulated by tumor stroma interaction signals, enabling non-CSCs to convert into CSCs.  \\
\\
    ...
\\
    Background: \{input\} \\
    Hypothesis: 
        \\
        \hline
        \bottomrule
    \end{tabular}
\end{table*}

\subsection{Prompts for Multi-agent Collaboration}
We present prompts for each role in multi-agent collaboration in Table~\ref{tab:appendix-multi-agent-role}, and prompts for environment settings in Table~\ref{tab:appendix-multi-agent-env}.

\begin{table*}
    \caption{
        Prompts for role design in multi-agent collaboration on hypothesis proposing task.
    }
    \label{tab:appendix-multi-agent-role}
    \centering
    \small
\begin{tabular}{p{\linewidth}}
\toprule
\hline
\underline{\textbf{\textsc{Prompts for role design in Multi-agent Collaboration}}} \\
\textbf{Analyst}:\\
You are the Analyst. Depending on the phase of the iteration, your role may slightly differ:\\
\\
- **Initial Phase**: Analyze the provided research background to distill its core components into pivotal keywords or topics. This will set the stage for the Engineer's search efforts.\\
- **Feedback Phase**: Based on feedback from the Critic, you might need to re-analyze the research background or provide additional insights to refine the search direction.\\
\\
In either case, ensure clarity and relevance in your analysis. Conclude by listing the identified keywords or topics or by providing revised insights.\\
\\
\\
\textbf{Engineer}:\\
You are the Engineer. Your task revolves around searching based on the received keywords or insights, and this can involve multiple iterations:\\
\\
- Plan your search strategies by crafting logical keyword combinations.\\
- Conduct systematic searches for each combination, meticulously gathering data and results.\\
- Refine your searches iteratively based on initial findings and any new insights from the Analyst.\\
\\
Your output should be comprehensive and organized. For each keyword combination:\\
\\
- **Title of Source**: Provide the title of the paper, article, or material you've found.\\
- **Abstract/Summary**: A brief summary or the abstract of the source.\\
- **Key Findings**: Highlight pivotal points or findings from the source that are relevant to the research background.\\
- **Implications**: If any, mention the implications or significance of the findings.\\
- **Relevant Quotes/Excerpts**: Extract direct quotes or sections that are particularly insightful.\\

Group your findings into individual "clues" based on themes or topics that emerge. This structure will provide the Scientist with detailed and organized data, enabling them to craft a robust hypothesis.\\
\\
Conclude by presenting the structured "clues" for each keyword combination.\\
\\
\\
\textbf{Scientist}:\\
You are the Scientist. Your task is to craft a hypothesis based on the Engineer's findings and the initial research background:\\
\\
- Derive a potential hypothesis that bridges the existing literature with new insights.\\
- Ensure the hypothesis is both innovative and scientifically grounded.\\
\\
Clearly state the proposed hypothesis, preparing it for evaluation by the Critic.\\
\\
\\
\textbf{Critic}:\\
You are the Critic, responsible for evaluating the collaborative endeavor. Scrutinize the Scientist's hypothesis in light of the `Research Background`. Gauge its novelty, coherence, and scientific validity. Should the hypothesis necessitate refinement:\\
\\
- Clearly articulate feedback, specifying areas needing improvement.\\
- Instruct the Analyst to either re-evaluate the `Research Background` or offer new insights to reshape the Engineer's subsequent search iteration.\\
\\
When the hypothesis aligns with expectations and meets the desired standards, present and approve it using the structured format:\\
\\
Final Answer:\\
(1) [First Point or Aspect of the Hypothesis]\\
(2) [Second Point or Aspect of the Hypothesis]\\
(3) [Third Point or Aspect of the Hypothesis]\\
...\\
\hline
        \bottomrule
    \end{tabular}
\end{table*}

\begin{table*}
    \caption{
        Prompts for environment setting in multi-agent collaboration.
    }
    \label{tab:appendix-multi-agent-env}
    \centering
    \small
\begin{tabular}{p{\linewidth}}
\toprule
\hline
\underline{\textbf{\textsc{Prompt for Environment setting in Multi-agent Collaboration.}}} \\
Topic Prompt for All Agents:\\
\\
You are part of a collaborative multi-agent system designed to propose a hypothesis based on a given research background. Each of you has a specific role:\\
\\
- **Analyst**: Analyzes the research background, distills its essence, and provides pivotal keywords or topics for further exploration.\\
- **Engineer**: Uses the keywords to plan and conduct systematic searches, meticulously gathering and organizing findings into detailed and structured "clues".\\
- **Scientist**: Crafts a potential hypothesis based on the organized findings and the original research background.\\
- **Critic**: Evaluates the hypothesis for its novelty, coherence, and scientific validity, providing feedback for refinement if necessary.\\
\\
Your collaboration is iterative. Based on feedback from the Critic, the process can loop back to the Analyst for refined insights, leading to new searches by the Engineer, and a refined hypothesis by the Scientist.\\
\\
Stay focused on your individual roles, collaborate effectively, and aim to derive a well-informed, novel hypothesis based on the research background provided.\\
\\
Research Background:\\
{background}\\
\\
Objective:\\
Using the research background and collaborative insights, the goal is to construct the most logical and scientifically robust hypothesis. Let's collaborate effectively to achieve this.
        \\
        \hline
        \bottomrule
    \end{tabular}
\end{table*}

\subsection{Prompts for ChatGPT Evaluation}
The instruction formats for prompting ChatGPT for evaluation on novelty, relevance, significance, and verifiability are displayed in Table~\ref{tab:appendix-eval-novelty}, Table~\ref{tab:appendix-eval-relevance}, Table~\ref{tab:appendix-eval-significance}, and Table~\ref{tab:appendix-eval-verifiability}, respectively.

\begin{table*}
    \caption{
        Prompts for ChatGPT evaluation on novelty metric.
    }
    \label{tab:appendix-eval-novelty}
    \centering
    \small
\begin{tabular}{p{\linewidth}}
\toprule
\hline
\underline{\textbf{\textsc{Prompt for ChatGPT evaluation on novelty metric.}}} \\
You are an expert in biomedicine.\\
Evaluate the novelty of the generated scientific hypothesis and the given background.\\
The score range should be 0 to 3. 0 means there's no novelty, which indicates that the hypothesis is a paraphrase of the background. 1 means there's slight novelty. 2 means there's moderate novelty. 3 means the hypothesis has strong novelty, which gives new insights beyond the background. Output is an integer.\\
\\
Please provide a step-by-step explanation supporting your score. \\
At the end of your response, clearly state the score in the format 'Score: [value]', where [value] can be 1, 2, or 3.\\
\\
Background: \{background\}\\
Generated scientific hypothesis: \{hypothesis\}
        \\
        \hline
        \bottomrule
    \end{tabular}
\end{table*}

\begin{table*}
    \caption{
        Prompts for ChatGPT evaluation on relevance metric.
    }
    \label{tab:appendix-eval-relevance}
    \centering
    \small
\begin{tabular}{p{\linewidth}}
\toprule
\hline
\underline{\textbf{\textsc{Prompt for ChatGPT evaluation on relevance metric.}}} \\
You are an expert in biomedicine.\\
Evaluate the relevance of the generated scientific hypothesis and the given background. \\
The score range should be 0 to 3. 0 means there's no relevance. 1 means there's slight relevance. 2 means there's moderate relevance. 3 means they are strongly related. Output is an integer.\\
\\
Please provide a step-by-step explanation supporting your score. \\
At the end of your response, clearly state the score in the format 'Score: [value]', where [value] can be 1, 2, or 3.\\
\\
Background: \{background\}\\
Generated scientific hypothesis: \{hypothesis\}
        \\
        \hline
        \bottomrule
    \end{tabular}
\end{table*}

\begin{table*}
    \caption{
        Prompts for ChatGPT evaluation on significance metric.
    }
    \label{tab:appendix-eval-significance}
    \centering
    \small
\begin{tabular}{p{\linewidth}}
\toprule
\hline
\underline{\textbf{\textsc{Prompt for ChatGPT evaluation on significance metric.}}} \\
You are an expert in biomedicine.\\
Evaluate the significance of the generated scientific hypothesis and the given background. \\
The score range should be 0 to 3. 0 means there's no significance, which indicates that the hypothesis is just a common knowledge. 1 means there's slight significance. 2 means there's moderate significance. 3 means the hypothesis has strong significance, which gives significant insights beyond the background. Output is an integer.\\
\\
Please provide a step-by-step explanation supporting your score. \\
At the end of your response, clearly state the score in the format 'Score: [value]', where [value] can be 1, 2, or 3.\\
\\
Background: \{background\}\\
Generated scientific hypothesis: \{hypothesis\}
        \\
        \hline
        \bottomrule
    \end{tabular}
\end{table*}

\begin{table*}
    \caption{
        Prompts for ChatGPT evaluation on verifiability metric.
    }
    \label{tab:appendix-eval-verifiability}
    \centering
    \small
\begin{tabular}{p{\linewidth}}
\toprule
\hline
\underline{\textbf{\textsc{Prompt for ChatGPT evaluation on verifiability metric.}}} \\
You are an expert in biomedicine.\\
Evaluate the verifiability of the generated scientific hypothesis and the given background. \\
The score range should be 0 to 3. 0 means there's no verifiability, which indicates that the hypothesis is not possible to be verified in future work. 1 means there's slight verifiability. 2 means there's moderate verifiability. 3 means the hypothesis has strong verifiability, which means the hypothesis is very likely to be verified in future work. Output is an integer.\\
\\
Please provide a step-by-step explanation supporting your score. \\
At the end of your response, clearly state the score in the format 'Score: [value]', where [value] can be 1, 2, or 3.\\
\\
Background: \{background\}\\
Generated scientific hypothesis: \{hypothesis\}
        \\
        \hline
        \bottomrule
    \end{tabular}
\end{table*}

\end{document}